\documentclass{article}

\usepackage[preprint, nonatbib]{timeseries_workshop}

\usepackage[utf8]{inputenc} % allow utf-8 input
\usepackage[T1]{fontenc}    % use 8-bit T1 fonts
\usepackage{hyperref}       % hyperlinks
\usepackage{url}            % simple URL typesetting
\usepackage{booktabs}       % professional-quality tables
\usepackage{amsfonts}       % blackboard math symbols
\usepackage{nicefrac}       % compact symbols for 1/2, etc.
\usepackage{microtype}      % microtypography
\usepackage{xcolor}         % colors
\usepackage{mathtools}
\usepackage{graphicx}
\usepackage{booktabs}

\title{Evaluating Time Series Foundation Models on Noisy Periodic Time Series}

\author{
  Syamantak Datta Gupta \\ %\thanks{Use footnote for providing further information
}

\begin{document}

\maketitle

\begin{abstract}
While recent advancements in foundation models have significantly impacted machine learning, rigorous tests on the performance of time series foundation models (TSFMs) remain largely underexplored. This paper presents an empirical study evaluating the zero-shot, long-horizon forecasting abilities of  several leading TSFMs over two synthetic datasets constituting noisy periodic time series. We assess model efficacy across different noise levels, underlying frequencies, and sampling rates.  As benchmarks for comparison, we choose two statistical techniques: a Fourier transform (FFT)-based approach and a linear autoregressive (AR) model. Our findings demonstrate that while for time series with bounded periods and higher sampling rates, TSFMs can match or outperform the statistical approaches, their forecasting abilities deteriorate with longer periods, higher noise levels, lower sampling rates and more complex shapes of the time series.
\end{abstract}

\section{Introduction}
In recent years, foundation models have ushered in dramatic breakthroughs to the field of machine learning\cite{foundation_models}, primarily due to their remarkable ability to excel at zero-shot and few-shot learning tasks. While  this revolution is the most remarkable in the context of language generation\cite{LLM_few_shots}, such models have gained prominence in other domains as well, including for time series data. However, while several foundation models for time series (TSFM) forecasting have been presented, rigorous comparative analysis of these models remains scarce, and our understanding of their performance beyond benchmark datasets is still limited. Unlike traditional statistical foundation models lack well-defined assumptions and theoretical guarantees and bounds, making it uncertain how they would perform forecasting previously unseen time series with arbitrary statistical properties. 

In this paper, we address some of these issues, empirically exploring  the efficacy of several TSFMs in zero-shot forecasting of noisy, periodic univariate time series over a long-horizon. The synthetic test data is generated as aggregates of sinusoids contaminated with noise, which are reasonable approximations for a large class of real-world time series.  We evaluate the impact of noise, underlying frequency, time series complexity, and sampling rate on model performance. The models considered are TimesFM, Lag-Llama, CHRONOS ("small" and "base"), and MOIRAI ("small" and "base").  We compare the TSFMs with two conventional statistical approaches: a simple, zero-shot approach that reconstructs the time series using a Fourier transform (FFT), and a linear autoregressive (AR) model. 

\textbf{Our contribution - we demonstrate the following:}
\begin{itemize}
 \item  TSFMs perform well on time series with shorter periods and smoother shapes. Under low noise and high sampling rates, some of them even outperform the statistical benchmarks.  
\item The forecasting abilities of  TSFMs decline sharply with very long periods, higher noise levels, lower sampling rates and more complicated shapes of the underlying time series. 
% \item We demonstrate that a majority of the TSFMs match the performance of the zero-shot FFT approach and they underperform compared to the linear AR model.
\end{itemize}

The rest of the paper is organized as follows. In section \ref{tsfm_intro}, we present a short introduction to time series foundation models and provide a brief description of the models we chose for this study. In section \ref{data}, we explain the methodology used for generating the synthetic data and the statistical forecasting methods we compare the TSFMs with. We present and discuss our results in section \ref{result}. Finally we summarize our findings and comment on future directions in section \ref{conclusion}.

\section{Time Series Foundation Models}
\label{tsfm_intro}
 Time Series Foundation Models (TSFMs)  aim to adapt the foundation model paradigm to the specific challenges of time series data. By leveraging extensive time series datasets, these models have the potential to achieve exceptional performance on a wide range of tasks, including zero-shot forecasting. 

Over the past few years, dozens of TSFMs have been presented, equipped with a diverse variety of architecture, pipelines and functionalities. Detailed surveys on these models can be found in \cite{tsfm_survey_1, tsfm_survey_2, tsfm_survey_3}. In this paper, we focus on a few of those: specifically, we refer to \cite{tsfm_survey_1}, to pick some of the state-of-the-art, transformer-based, self-supervised generative TSFMs, that are capable of making zero-shot long-horizon forecasts on univariate time-series. 

The models we consider are as follows: \textbf{Lag-Llama}\cite{LagLLama} - A decoder-only model that uses the architecture of LLaMA\cite{llama_2} as a backbone.
\textbf{CHRONOS}\cite{chronos} - A TSFM that leverages the architecture of the encoder-decoder model of T5\cite{T5}.
\textbf{MOIRAI} \cite{moirai} - A TSFM based on masked encoder-based universal forecasting transformer.
\textbf{TimesFM}\cite{timesFM} - A patch-based decoder-only model, trained from scratch on time series data. 
%The first two utilize the architecture of pre-trained language models, while the latter two present novel architectures. 
For the MOIRAI and CHRONOS models, we use both the "small" and the "base" versions. While these models represent only a fraction of the diversity of TSFMs currently available, they are among the best performing of their kind as demonstrated in the respective papers. 

% \begin{enumerate}
%     \item \textbf{Lag-Llama}\cite{LagLLama} - A decoder-only model that uses the architecture of LLaMA\cite{llama_2} as a backbone
%     \item \textbf{TimesFM}\cite{timesFM} - A decoder-only model, trained from scratch exclusively on time series data
%     \item \textbf{Chronos}\cite{chronos} - A TSFM that leverages the architecture of the encoder-decoder T5\cite{T5} model
%     \item \textbf{MoirAI} \cite{moirai} - A TSFM based on masked encoder-based universal forecasting transformer
% \end{enumerate}

\section{Synthetic Data Generation and FFT-benchmark}
\label{data}
% In this section we briefly discuss the theoretical considerations that motivated our experiment and the statistical benchmarks that follow. 
%\subsection{The Fourier Series}
% The sampling theorem states that any  signal $X(t)$ with a maximum frequency component $f_{max}(t)$, can be uniquely reconstructed from discrete observations if the sampling rate is greater than the twice of  $f_{max}(t)$. Formally, let the continuous-time signal $X(t)$ be represented as follows:
% \begin{equation}
%  X(t) = A_0 + \sum_{n=1}^N A_n \sin \left(2 \pi f_n t + \phi_n\right),
% \end{equation}
% where $f_n$, $A_n$, $\phi_n$ are respectively the frequencies, amplitudes and phases (in radians) of the components, and let $f_{max} = \max \{ f_n \}$. If $X(t)$ is uniformly sampled at intervals of $\tau$, then it can be fully reconstructed from those samples as long as $\frac{1}{\tau} > 2f_{max}$. When $f_{\max}< \infty$,  $X(t)$ is called a band-limited signal. The cut-off frequency $ 2f_{max}$ is called the Nyquist rate. The sampling theorem is a fundamental result in signal processing and forms the basis of  all  forms of digital communication. 
\subsection{Synthetic Data Generation}

A wide variety of real-world data can be approximated by an aggregate of sinusoids of varying frequency, mixed with additive noise. These include meteorological data, radio and audio waves, EEG and ECG sequences, as well as electromagnetic signals of other kinds. Often a discrete-time, sampled version of the original continuous signal is available. Because of their relatively stable statistical properties over long time horizons, they make for an interesting case study for multi-step forecasting models. With this motivation, the synthetic data is generated as weighted sums of sinusoids with additive Gaussian noise. First, a continuous-time version is produced as: 

\begin{equation}
    X(t) = \sum_{n=1}^N A_n \sin \left( 2 \pi f_n t + \phi_n\right) + w(t), \label{mixed_sins}
\end{equation}
where $w(t)$ is zero-mean, uncorrelated Gaussian noise $\sim \mathcal{N}(0, \sigma^2)$. As long as the frequencies $f_n$ are rational, the noise-free part of $X(t)$ is periodic. Following the Nyquist theorem, to avoid aliasing, $X(t)$ is uniformly sampled at the rate of $f^* > 2f_{\max}$, where $f_{\max} = \max \{f_n\} $, to obtain the test time series. We consider two versions of (\ref{mixed_sins}) to generate two sets of test data, that we call set A and set B. In set A, the higher frequencies are generated as multiple of the lowest frequency $f_1$, so that $f_n = nf_1$, with $f_1$ being an integer. In this case the time series takes the form:

\begin{equation}\label{eq:periodic}
    X(t) = \sum_{n=1}^N A_n \sin \left(2 \pi nf_1t + \phi_n\right) + w(t).
\end{equation}
It follows from Fourier's theorem that any continuous periodic series can be approximated by a weighted sum of its harmonics, so the form in (\ref{eq:periodic}) represents a noisy periodic signal with frequency $f_1$. In contrast, in set B, the time series assume the more general form of (\ref{mixed_sins}): the frequencies $f_n$ are allowed to assume any positive rational value within a range. By construction, the underlying noise-free part of $X(t)$ in set A has a time period $1/f_1 \leq 1$ (since $f_1$ is chosen to be an integer). For set B,  this period is given by the least common multiplier of $b_1$, $b_2$, $\dots$, $b_N$, where $f_n = \frac{a_n}{b_n}$ in fractional form. This is unbounded, and for randomly chosen $f_n$, can be arbitrarily large. %In other words, the time series in B have long memory. 

% To generate a diverse collection of time series for testing, we vary the number of components N, the noise variance $\sigma^2$ and the sampling rate $\tau$. 
For both sets, we vary the number of components N in  [1, 2, 3, 5, 8, 12, 20], and randomly generate 20 sets of amplitudes $A_n$ and frequencies $f_n$ for each N. The phase $\phi_n$ is kept 0. For set A, the principal frequencies $f_1$ are randomly chosen, whereas for set B, all the frequencies are randomly picked.  For each amplitude-frequency combination, we vary $\sigma^2$ so that the signal-to-noise ratio (SNR) in dB takes values of 2, 5, 10, 15, 20 and 30. Finally the sampling rate $f^*$ is also varied as the following multiples of $f_{\max}$: [2.1,  2.5, 3, 5, 10, 20], where $f_{\max} = \max \{f_n\} $. With these choices, for sets A and B we generate 5,040 time series each, each with 576 observations. We consider a context length of 512 and a forecast horizon of 64 for each model tested. %These numbers are chosen based on context and forecast length recommendations of some of the TSFMs evaluated. 

\subsection{Statistical Approaches for Comparison: FFT and Linear AR}
As an elementary "zero-shot" alternative, we use Fourier decomposition. First we apply a Hann\cite{Hanning, prabhu2014window} window on the input sequence to reduce the impacts of spectral leakage.  Then, for a context length \( T_1 \) and forecast horizon \( T_2 \), we first obtain the FFT\cite{FFT_1965} of the windowed \( T_1 \) observations. The FFT is filtered by clipping all values below a pre-defined threshold (We consider a flat threshold of 20\% of the peak FFT value). Estimates of amplitude, frequency and phase are recovered from this filtered FFT, which are then used to reconstruct the original time series over the combined context and forecast period (\( T_1 + T_2 \)) using equation (\ref{mixed_sins}). The last \( T_2 \) values of this sequence are our forecast. Notably, with no training or model fitting involved, this is in effect a "zero-shot" method. 

For the AR model, we choose the optimal model order $p$ by a grid search using the Akaike Information Criterion \cite{akaike}. The series $X(t)$ is estimated as a linear combination of its $p$ past values:
\begin{equation}
\hat{X}(t) = \sum_{k=1}^{p} A_k X(t-k).
\end{equation}
% The corresponding AR parameters $A_k$ can be computed using the sample autocovariances of  $X(t)$. 
 No additional regularization is done to limit the impact of overfitting, so the model remains linear.

\section{Results}\label{result}

The average and median mean squared forecasting errors (MSE)  are presented in Table \ref{table_64}.  On set A, the CHRONOS models lead other TSFMs and the FFT-based approach by a substantial margin on both metrics, falling only slightly short of the linear AR model. TimesFM finishes second among the TSFMs, doing similar to the FFT. On set B, the linear AR model outperforms all the TSFMs, and the FFT-based model finishes second in terms of average MSE. Among the TSFMs, the CHRONOS models and TimesFM outperform Lag-Llama and the MOIRAI models. In the following figures, we  further investigate these results, using the suffixes "\_b" and "\_s" to indicate the "base" and "small" versions of a TSFM. 

\begin{table}[ht!]
\caption{Mean and Median of forecasting MSE for sets A and B: \\  Metrics for the best performing model in bold, best performing TSFM underlined}
  \label{table_64}
    \centering
    \resizebox{\textwidth}{!}{ % Rescale the table to fit within the text width
    \begin{tabular}{lrrrrrrrr}
        \toprule
        & MOIRAI-base & MOIRAI-small & Lag-Llama & CHRONOS-base & CHRONOS-small & TimesFM & FFT & AR \\
        \midrule
        	
        set A, mean &30.33  &31.32  &30.17  & \underline{11.62} &  13.94 & 23.60 & 23.54 & \textbf{10.83} \\
        set A, median &21.09 & 21.84 &	21.13 &	\underline{2.18} &	2.21 &	12.99	& 15.44 &	\textbf{1.85} \\
        set B, mean & 42.31  &43.01  & 42.75 & 35.02 & 36.03 & \underline{33.32}	 & 32.42 & \textbf{13.32}\\
        set B, median & 27.83	& 28.34 &	27.61 &	\underline{14.55} &	16.14 &	19.80 &	19.62	& \textbf{2.39}\\
        % Set A, rank &5.66	 & 5.76 &	5.26 &	\textbf{1.87} &	2.25 &	3.65 &	3.55\\
        % Set B, rank & 5.39	 & 5.43 &	5.06 &	2.97 &	3.18 &	3.1 & \textbf{2.87}  \\
        \bottomrule
    \end{tabular}
     }
\end{table}

% The key difference between set A and set B is the effective periodicity of the underlying clean (noise-free) signal. By construction, the time period of all the series in set A is bounded by 1, whereas in set B, as the constituent sinusoids have arbitrary frequencies, the time periods are unbounded. As a result, the sequences have more complex waveforms, especially as the number of constituent sinusoidal components N goes high.

\begin{figure}[ht!]
    \centering
    \begin{tabular}{ccc}
        \includegraphics[width=0.3\textwidth]{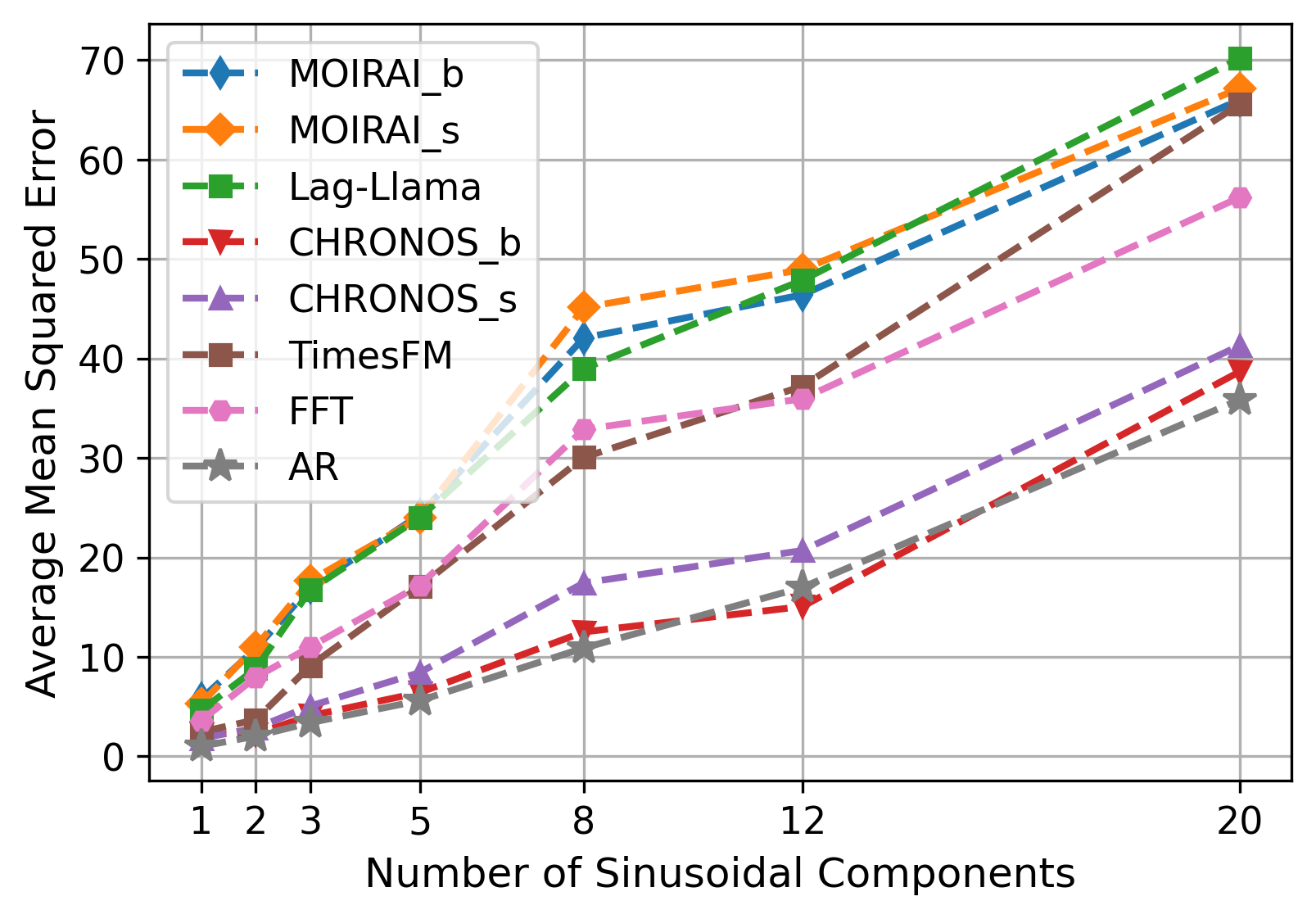} &
        \includegraphics[width=0.3\textwidth]{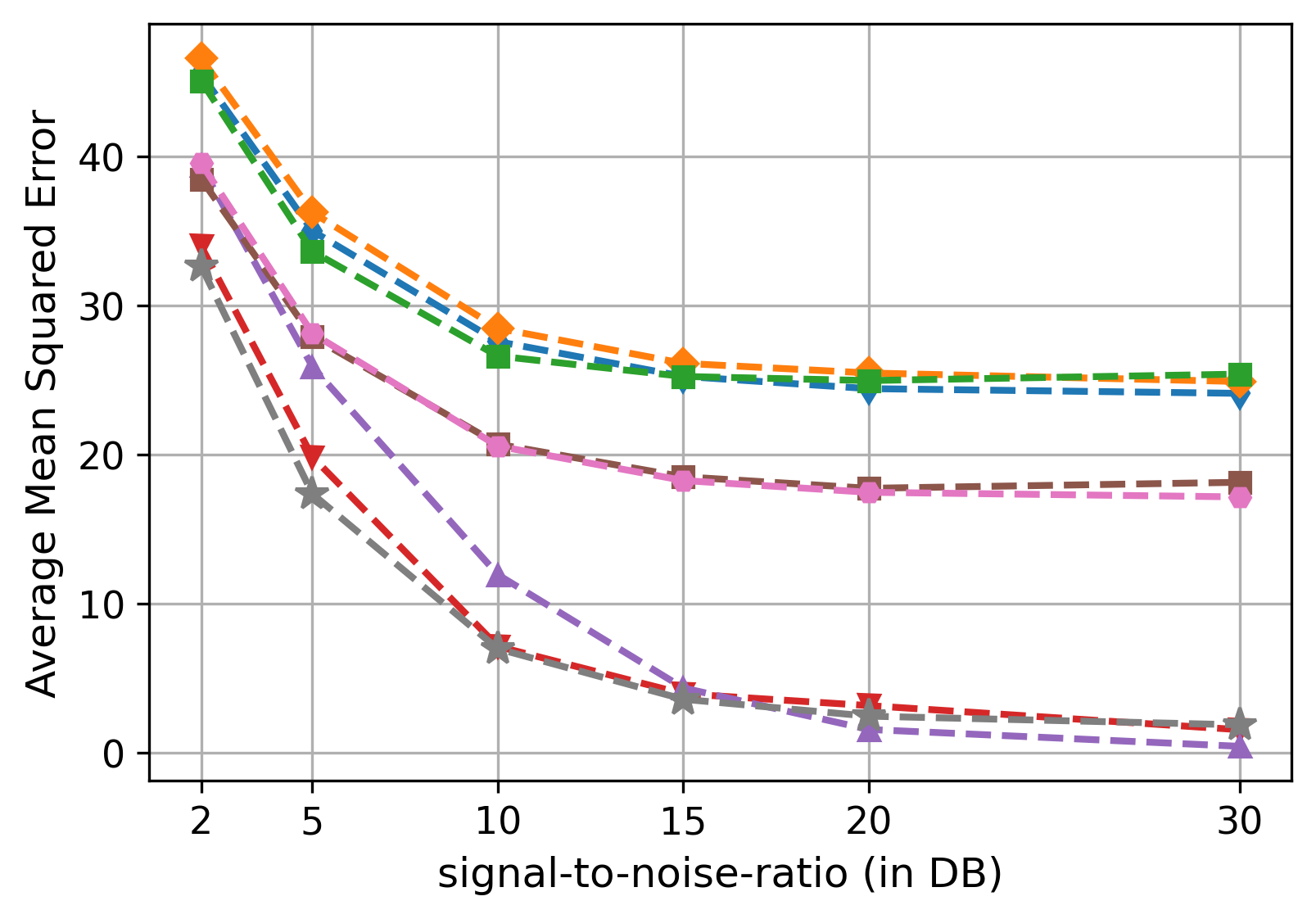} &
        \includegraphics[width=0.3\textwidth]{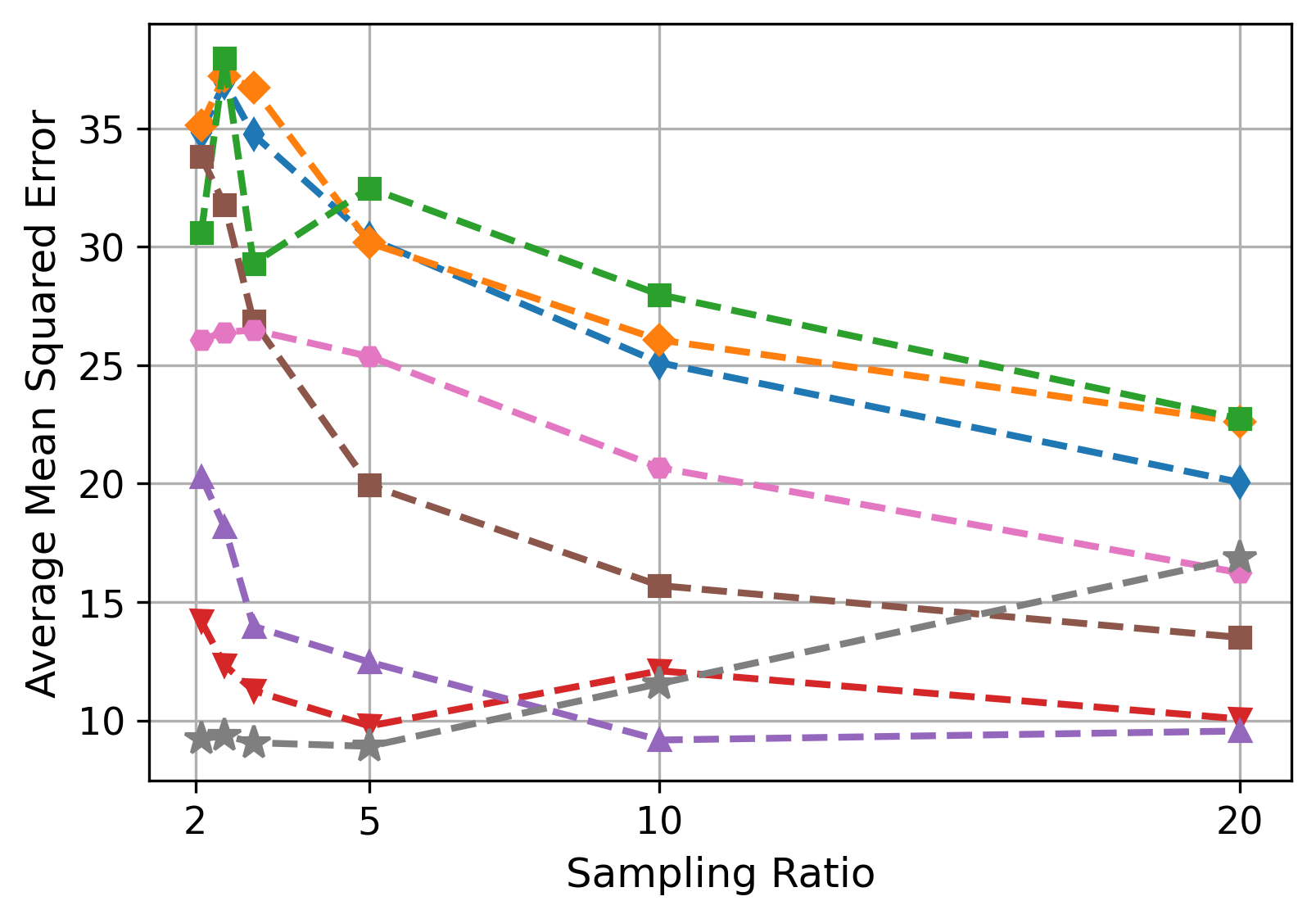} \\
        %(a) & (b) & (c) \\
        \includegraphics[width=0.3\textwidth]{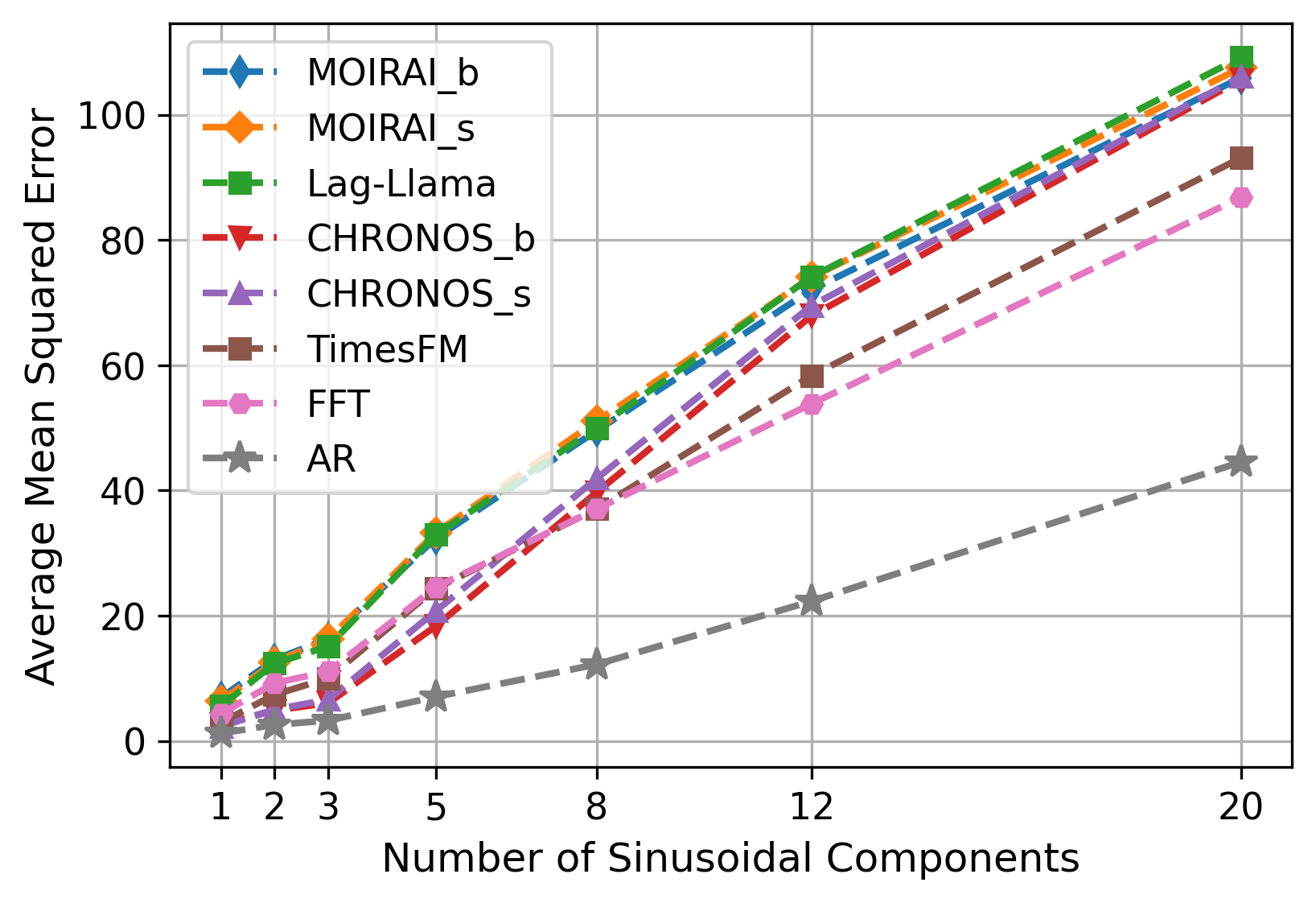} &
        \includegraphics[width=0.3\textwidth]{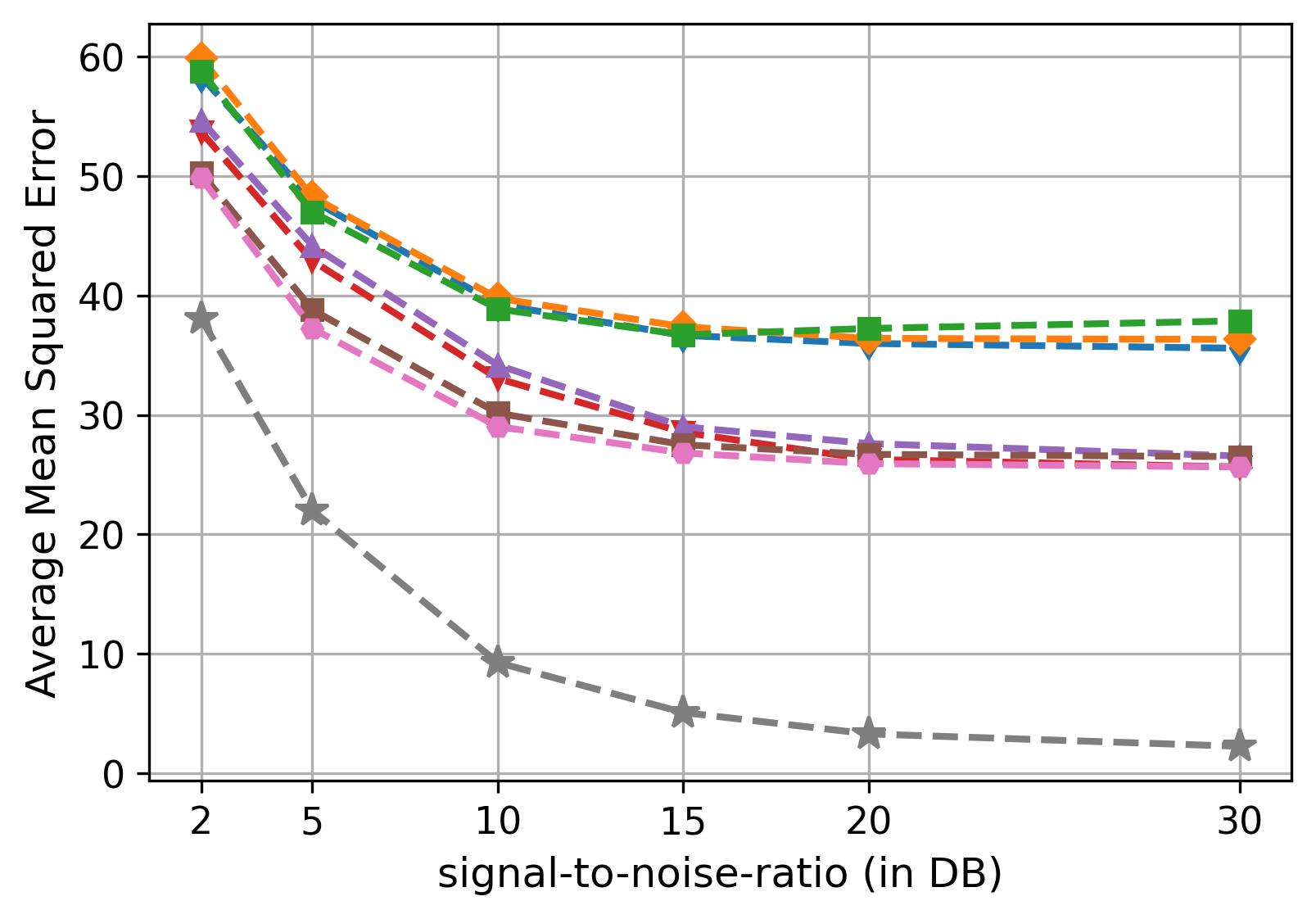} &
        \includegraphics[width=0.3\textwidth]{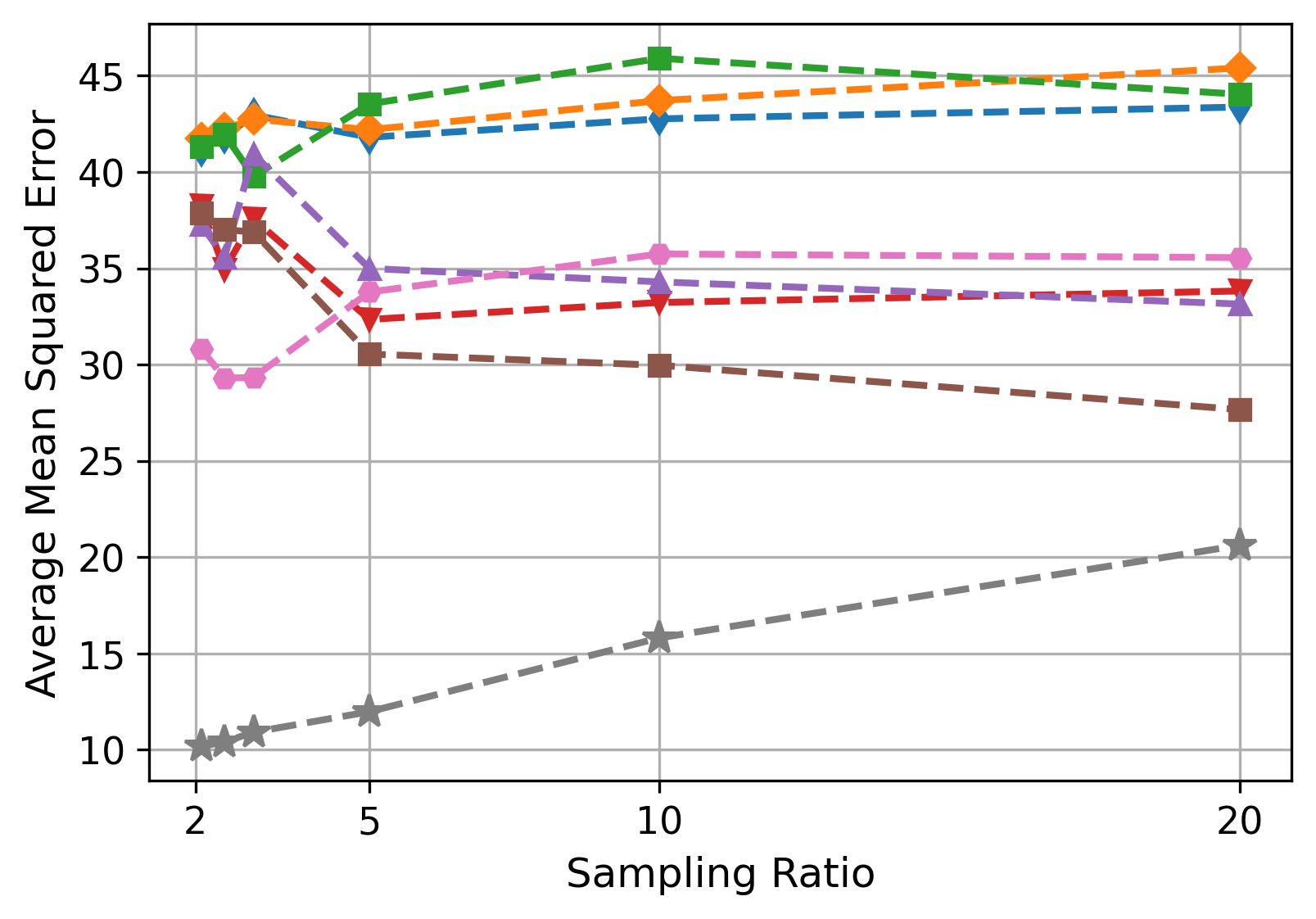} \\
       % (d) & (e) & (f)
    \end{tabular}
    \caption{Average Mean-Squared Error as a function of number of sinusoidal components, signal-to-noise ratio and sampling ratio. Top row represents Set A and bottom row represents Set B}
    \label{fig:MSE_1}
\end{figure}

In Figure \ref{fig:MSE_1}, the average MSEs are broken down by the number of sinusoidal components N, signal-to-noise ratio (SNR) and sampling ratio. On set A, while the AR model performs marginally better at higher N, lower SNRs and lower sampling ratios, the CHRONOS models follow closely and even outperform the former at higher sampling ratios, where the AR model overfits.  However, on set B, the AR model demonstrates significant superiority, which further increases with higher N, lower noise and lower sampling ratios. The FFT, too, performs slightly better than the TSFMs on set B. 

\begin{figure}[ht!]
    \centering
    \begin{tabular}{cc}
        \includegraphics[width=0.45\textwidth]{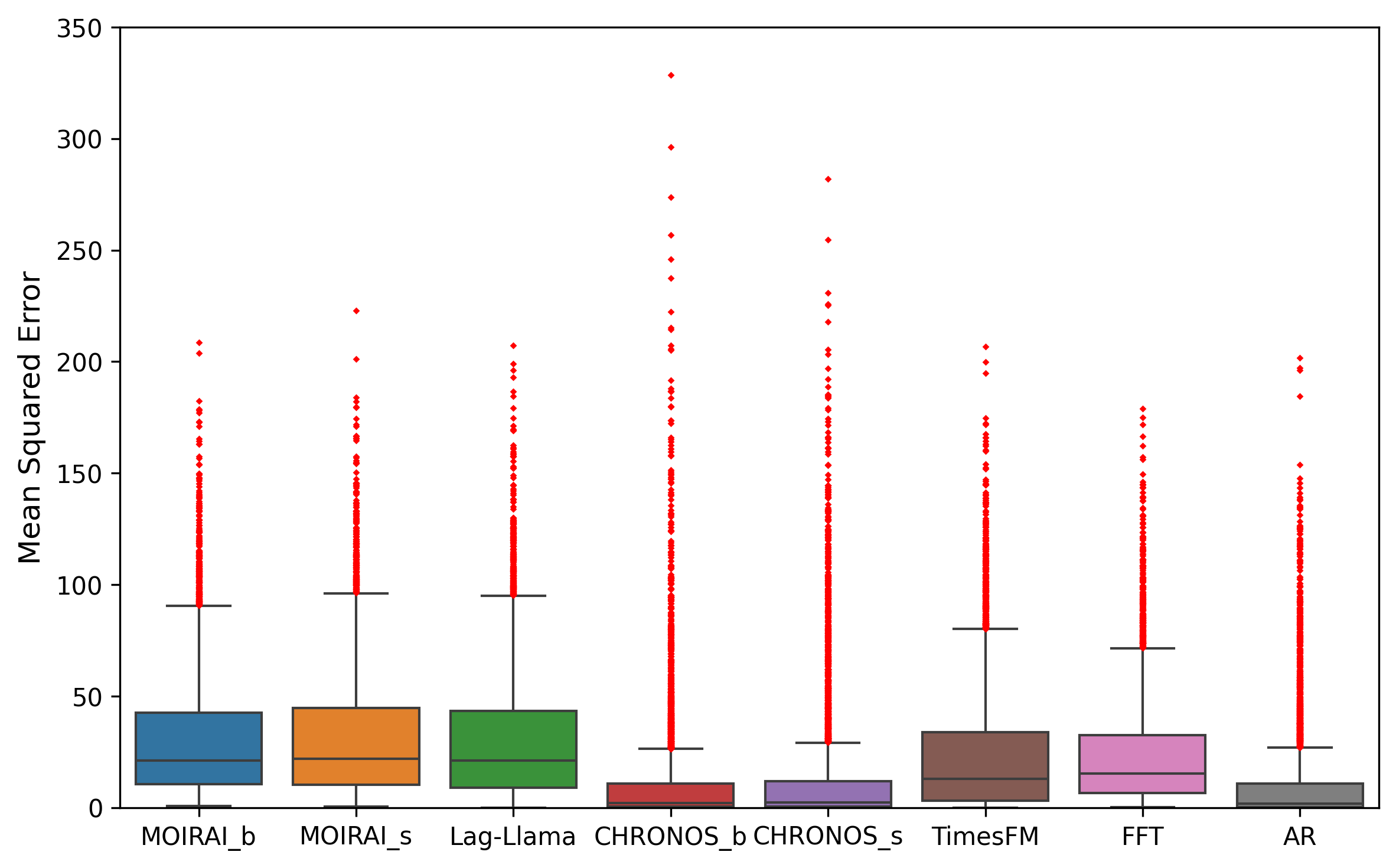} &
        \includegraphics[width=0.45\textwidth]{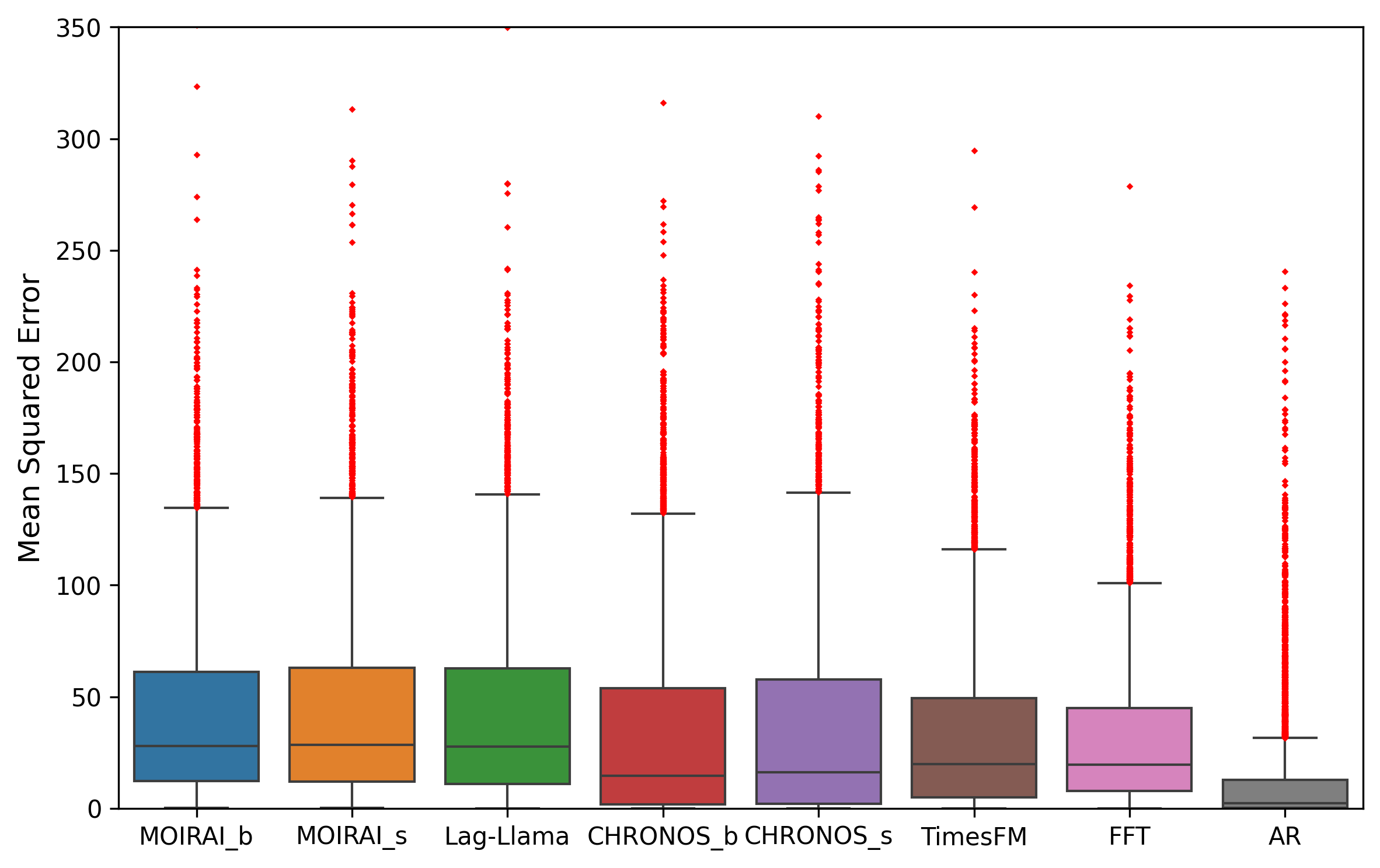} \\
        % (a) Caption for the first figure. & (b) Caption for the second figure. \\
    \end{tabular}
    \caption{Boxplots of mean-squared errors and outliers, left: set A, right: set B}
    \label{fig:boxplot}
\end{figure}

The boxplots of Figure \ref{fig:boxplot} represent distributions of the estimated MSEs for different models. For set A, the CHRONOS models have significantly short inter-quartile ranges, comparable only to that of the AR model. The inter-quartile ranges for all approaches except the AR model increase substantially from set A to set B. This increase is the most remarkable for the CHRONOS models where it no longer maintains a clear superiority over the other TSFMs. 

The distinguishing attribute of the time series in set B are their arbitrarily large periods (a median time-period of $10^9$ s, compared to $0.1$ s in set A),  so the 512-step context windows often miss a full cycle of observations. While the AR model and FFT can still extract useful information using lagged observations and frequency components, TSFMs struggle with this obstacle. These challenges are further exacerbated with higher N, lower sampling rates, and lower SNR.

A higher N adds more complexity to the shape (i.e, a higher number of peaks within a time period) of the time series, while a lower sampling ratio makes the series coarser, with more zigzag patterns. Both these conditions tend to impact TSFMs more adversely than they do so for the statistical methods. Contrarily, TSFMs perform better with smoother time series with fewer peaks and troughs.

\section{Conclusion}\label{conclusion}

In this paper, we evaluate zero-shot, long-horizon forecasting abilities of several TSFMs on noisy periodic time series. We show that for time series with bounded periods, TSFMs achieve strong performance, and the best TSFMs can outperform both an un-regularized linear AR model and an elementary FFT-based method when sampling rates are sufficiently large. However, when the periods of the series are arbitrarily large, TSFMs struggle to match the performance of the linear AR model and deteriorate further with increasing number of components, lower sampling rates and lower SNR. The CHRONOS models offer the best results among the TSFMs, followed by TimesFM. 

While by no means exhaustive, our results indicate existing limitations and potentials for improvement for the zero-shot forecasting abilities of TSFMs we studied. Fine-tuning these models on datasets like set B, for instance, could improve their performance and robustness. These findings also underscore the necessity for extensive evaluation of TSFMs over more diverse datasets, both synthetic and real-world, to identify specific attributes of the data where these models underperform or excel.

\medskip

{
\small

\bibliographystyle{IEEEtran}
\bibliography{tsfm}

% [1] @article{bommasani2021opportunities,
%   title={On the opportunities and risks of foundation models},
%   author={Bommasani, Rishi and Hudson, Drew A and Adeli, Ehsan and Altman, Russ and Arora, Simran and von Arx, Sydney and Bernstein, Michael S and Bohg, Jeannette and Bosselut, Antoine and Brunskill, Emma and others},
%   journal={arXiv preprint arXiv:2108.07258},
%   year={2021}
% }.

% [2] Bower, J.M.\ \& Beeman, D.\ (1995) {\it The Book of GENESIS: Exploring
%   Realistic Neural Models with the GEneral NEural SImulation System.}  New York:
% TELOS/Springer--Verlag.

% [3] Hasselmo, M.E., Schnell, E.\ \& Barkai, E.\ (1995) Dynamics of learning and
% recall at excitatory recurrent synapses and cholinergic modulation in rat
% hippocampal region CA3. {\it Journal of Neuroscience} {\bf 15}(7):5249-5262.
% }

%%%%%%%%%%%%%%%%%%%%%%%%%%%%%%%%%%%%%%%%%%%%%%%%%%%%%%%%%%%%

%%%%%%%%%%%%%%%%%%%%%%%%%%%%%%%%%%%%%%%%%%%%%%%%%%%%%%%%%%%%

\newpage
\appendix
\section{Appendix}

\begin{figure}[ht!]
    \centering
    \begin{tabular}{cc}
        \includegraphics[width=0.45\textwidth]{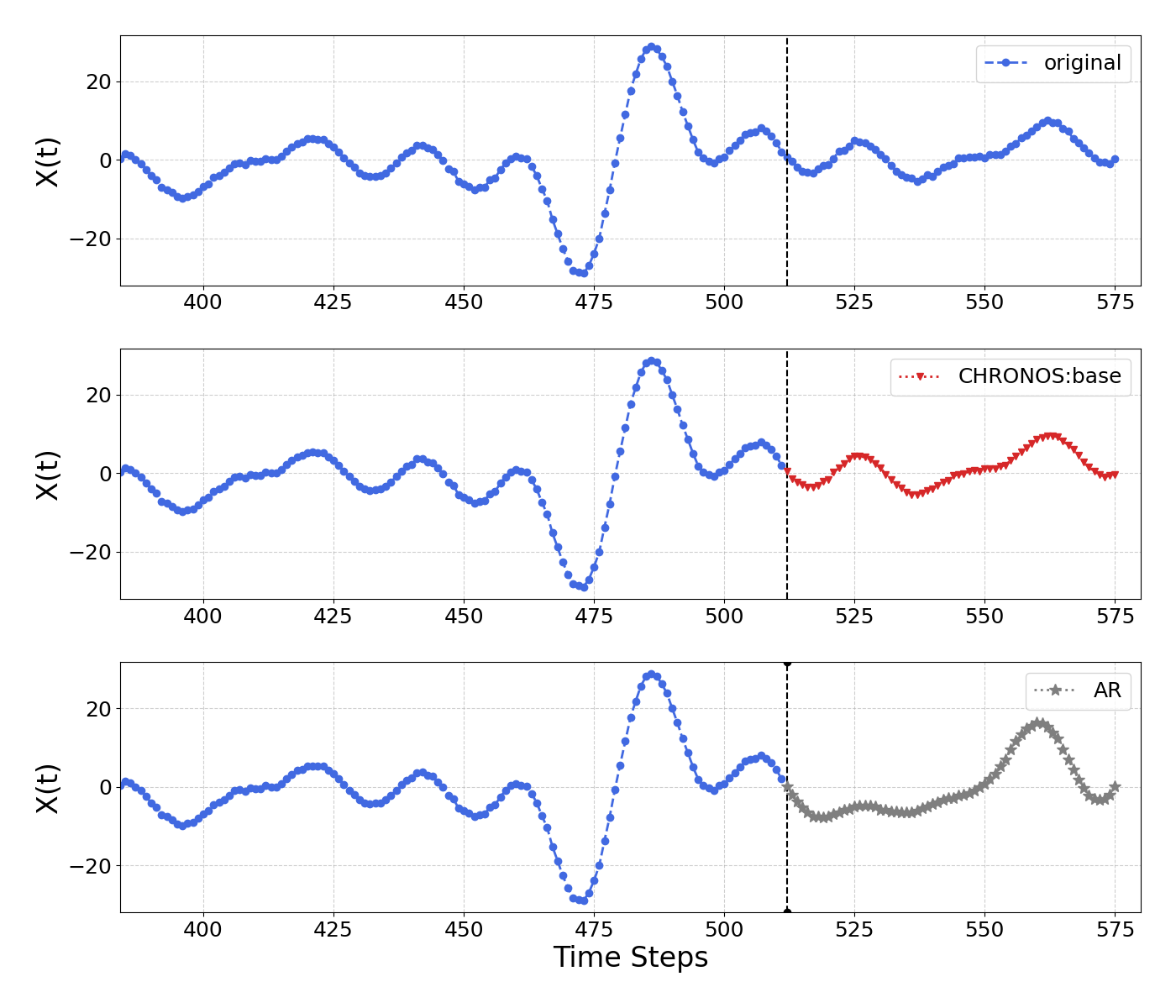} &
        \includegraphics[width=0.45\textwidth]{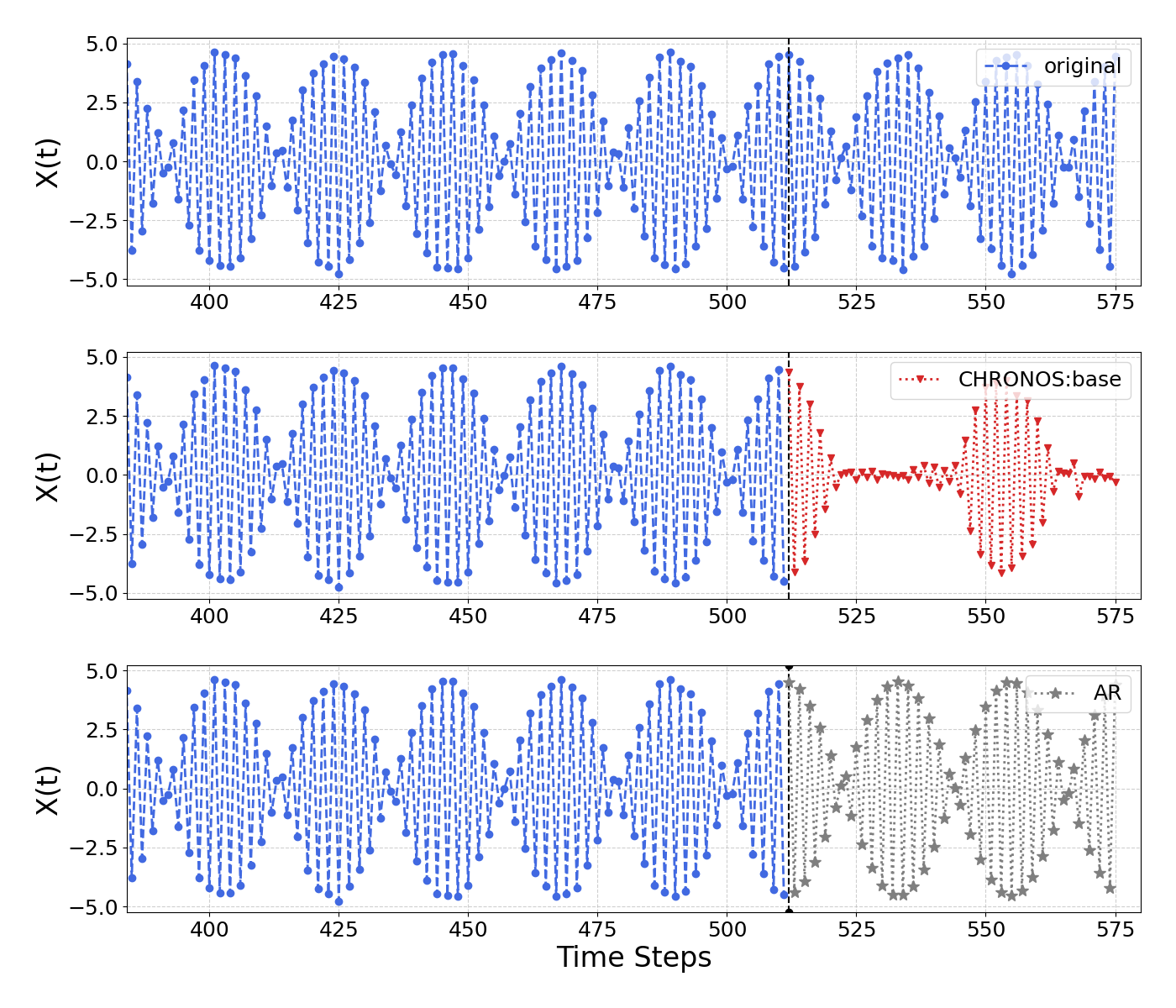} \\
        % (a) Caption for the first figure. & (b) Caption for the second figure. \\
    \end{tabular}
    \caption{Example time series and forecasts (context window truncated) CHRONOS-base and AR, left: set A, right: set B. The CHRONOS model beats AR for the smoother time series in set A but misses the pattern for the zig-zag series in set B, resulting from a lower sampling rate}
    \label{fig:examples}
\end{figure}
\begin{figure}[ht!]
    \centering
    \begin{tabular}{ccc}
        \includegraphics[width=0.3\textwidth]{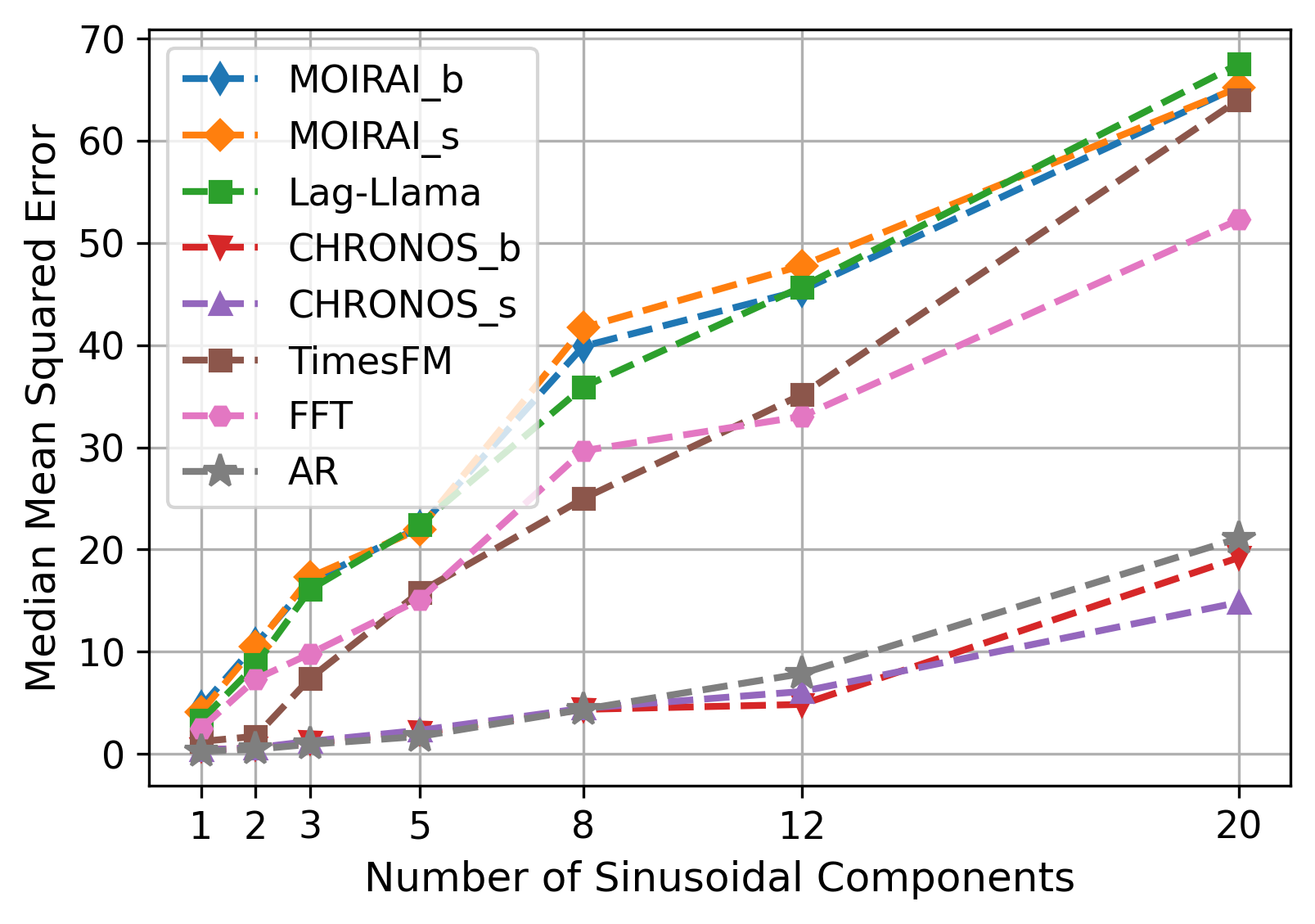} &
        \includegraphics[width=0.3\textwidth]{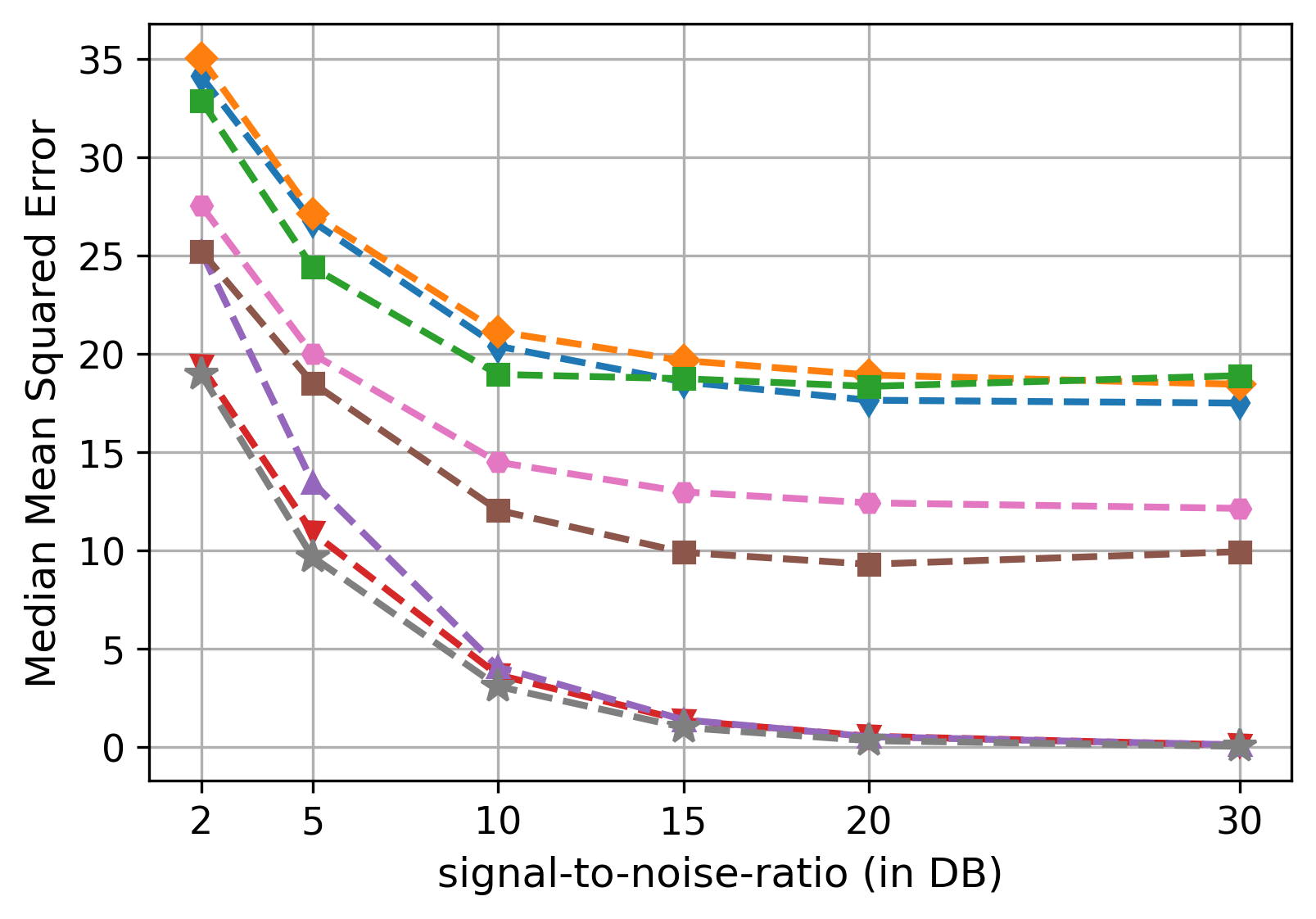} &
        \includegraphics[width=0.3\textwidth]{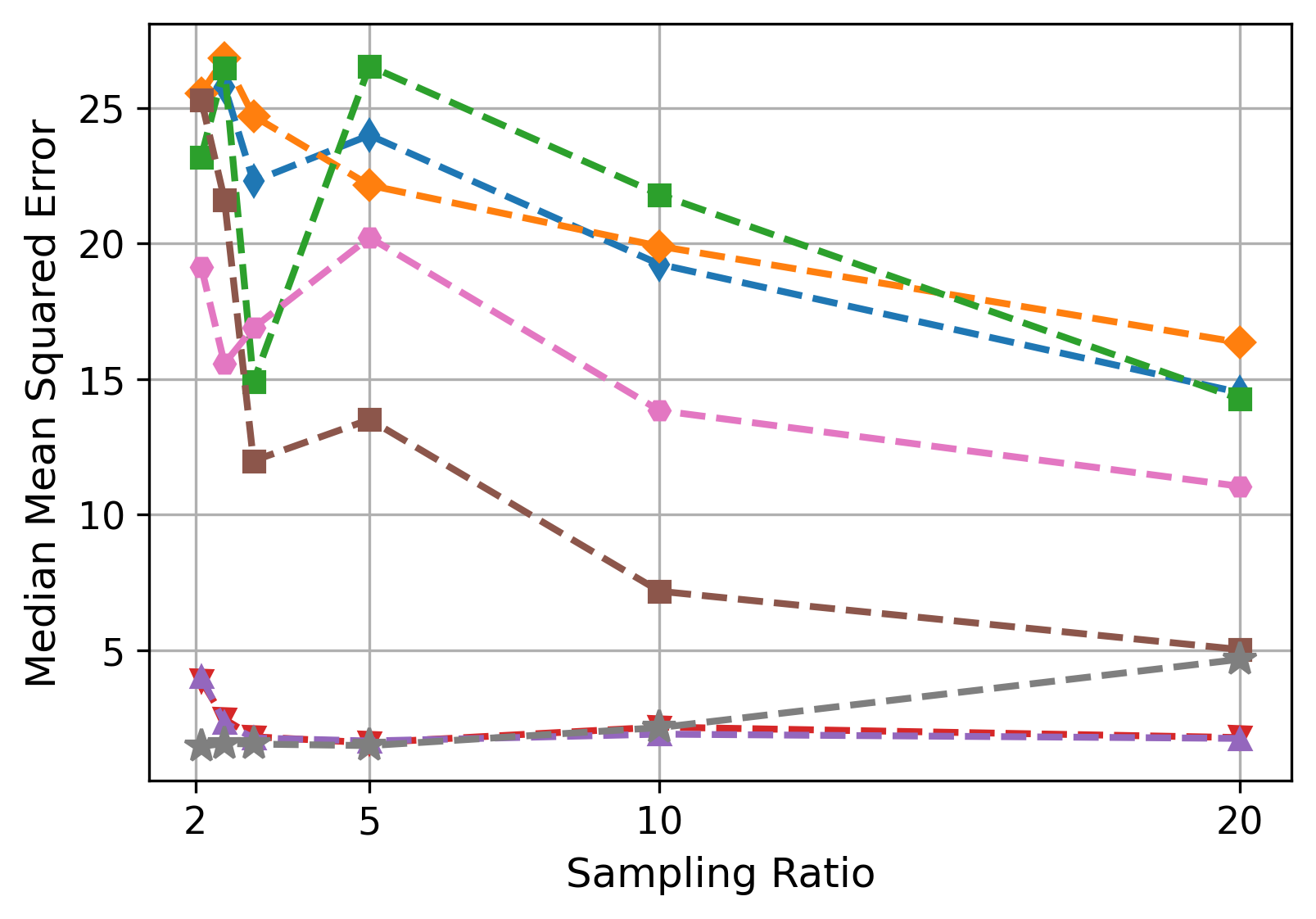} \\
        %(a) & (b) & (c) \\
        \includegraphics[width=0.3\textwidth]{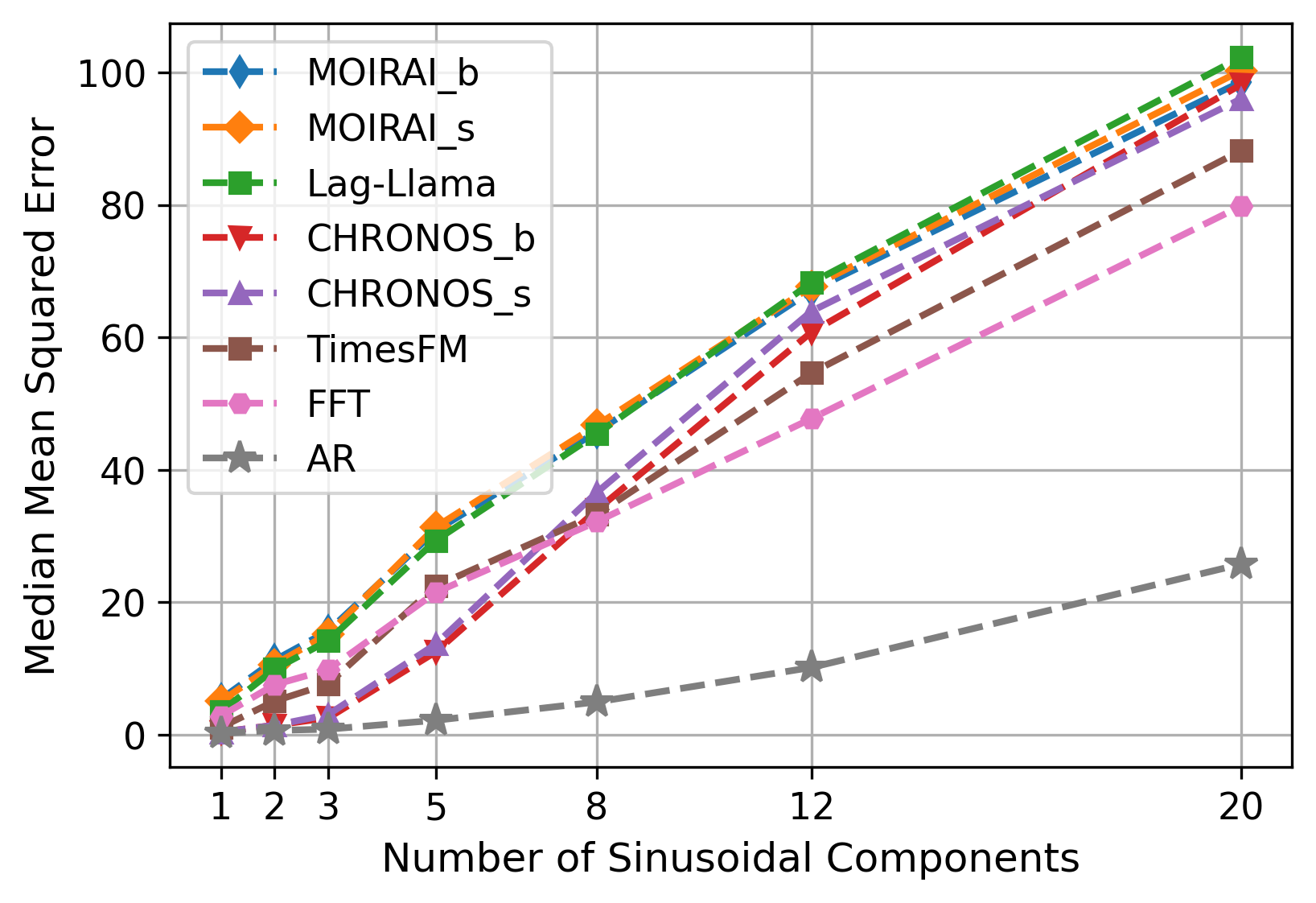} &
        \includegraphics[width=0.3\textwidth]{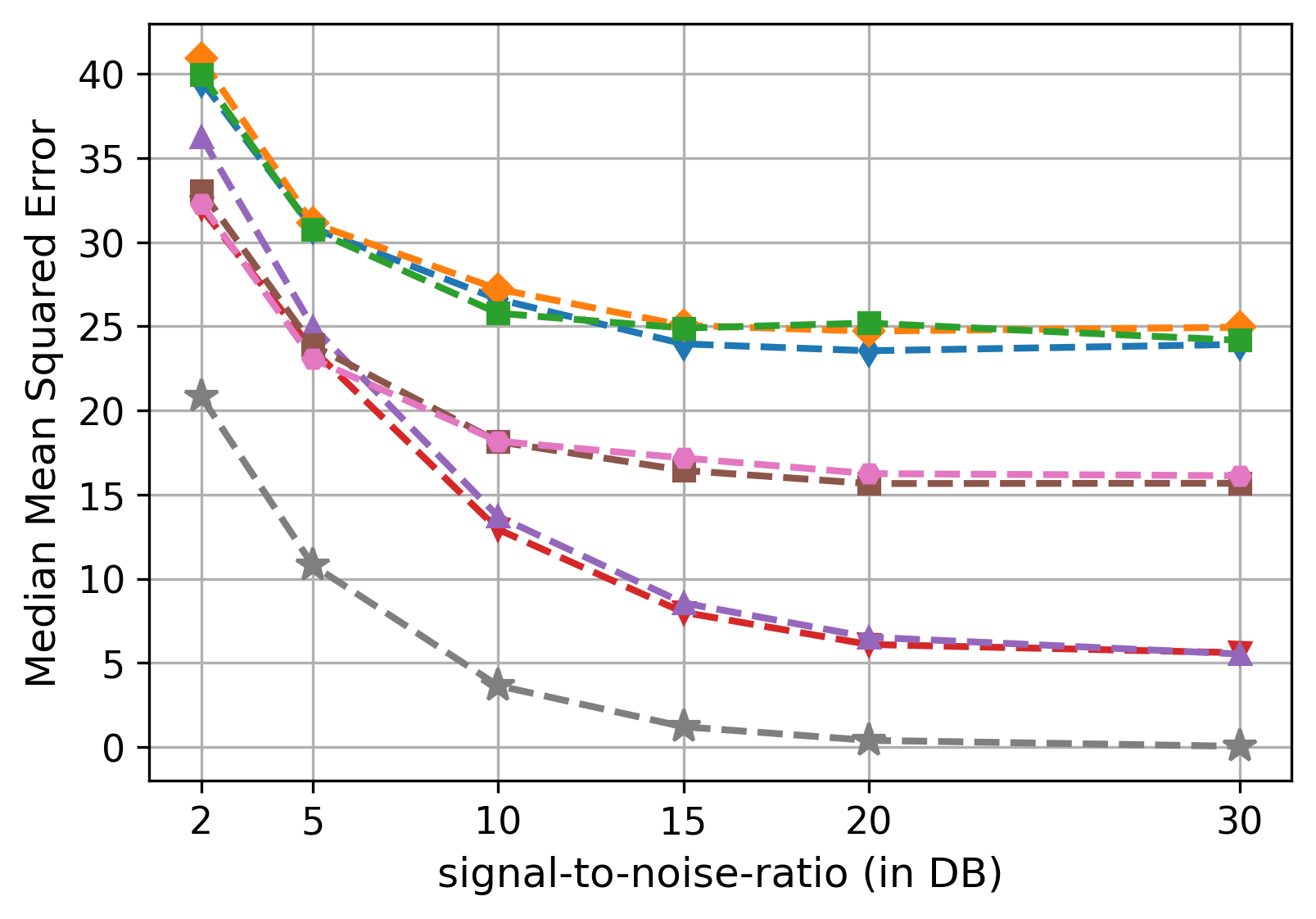} &
        \includegraphics[width=0.3\textwidth]{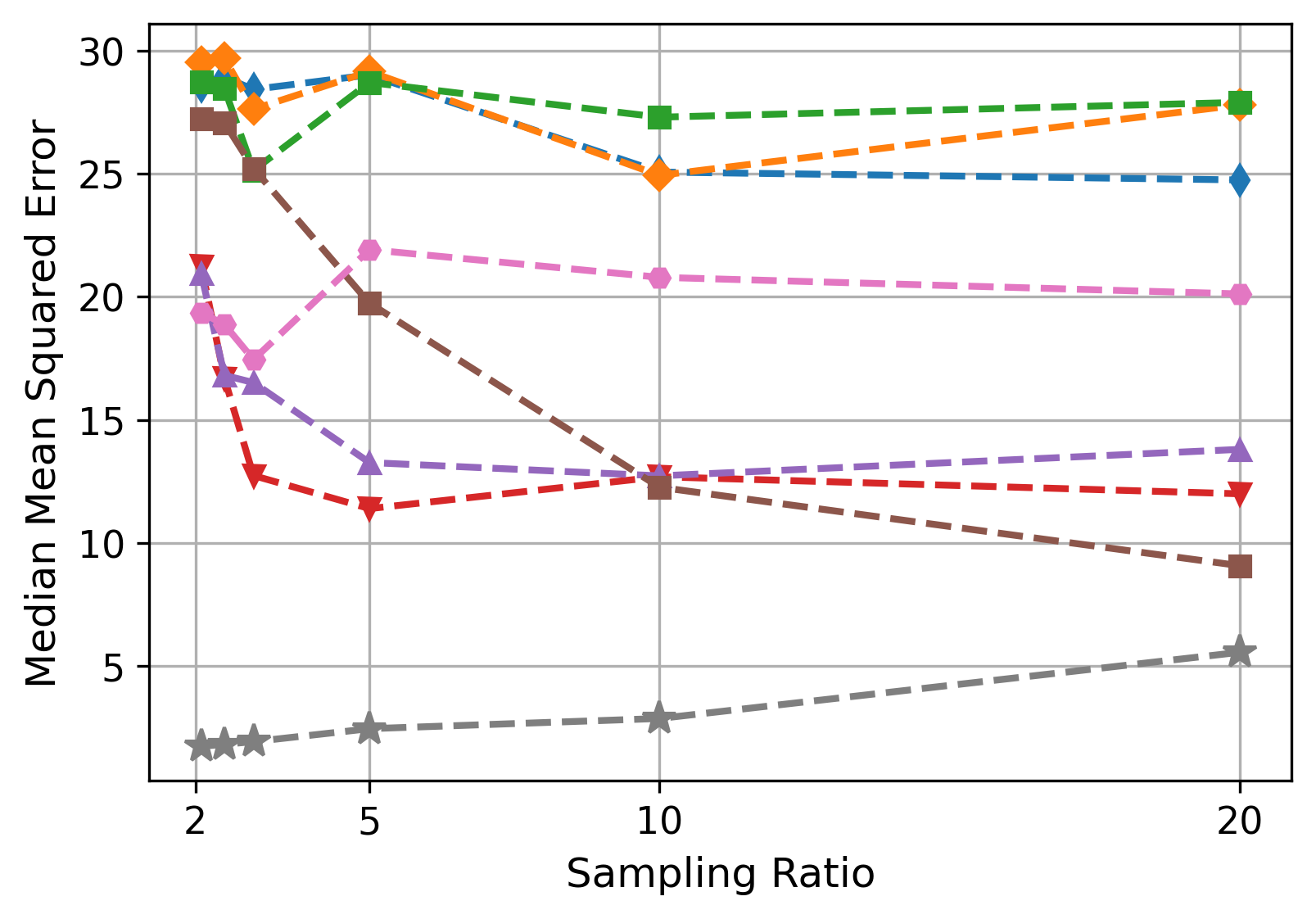} \\
       % (d) & (e) & (f)
    \end{tabular}
    \caption{Median Mean-Squared Error as a function of number of sinusoidal components, signal-to-noise ratio and sampling ratio. Top row represents Set A and bottom row represents Set B}
    \label{fig:MSE_median}
\end{figure}

\begin{figure}[ht!]
    \centering
    \begin{tabular}{ccc}
        \includegraphics[width=0.3\textwidth]{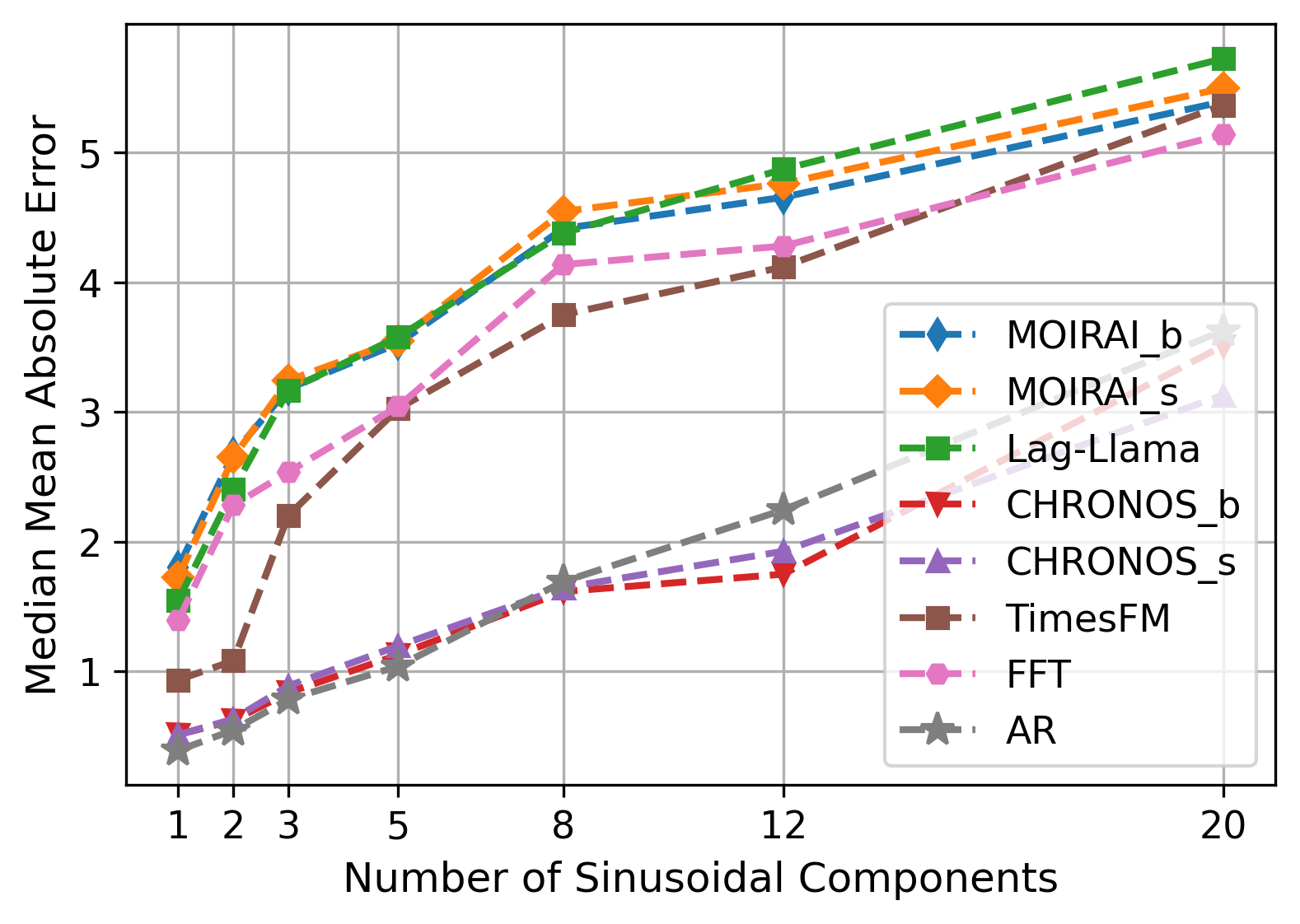} &
        \includegraphics[width=0.3\textwidth]{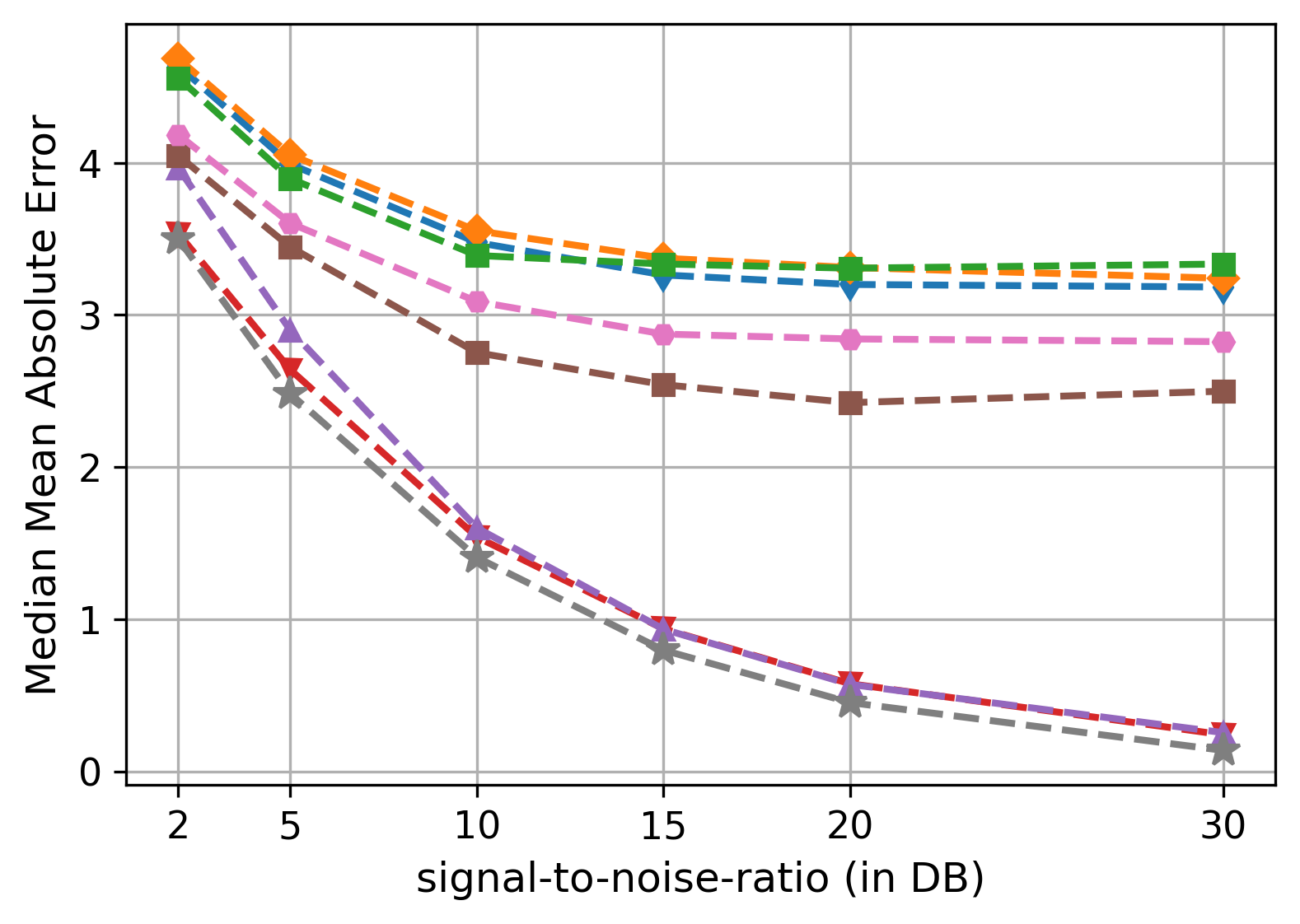} &
        \includegraphics[width=0.3\textwidth]{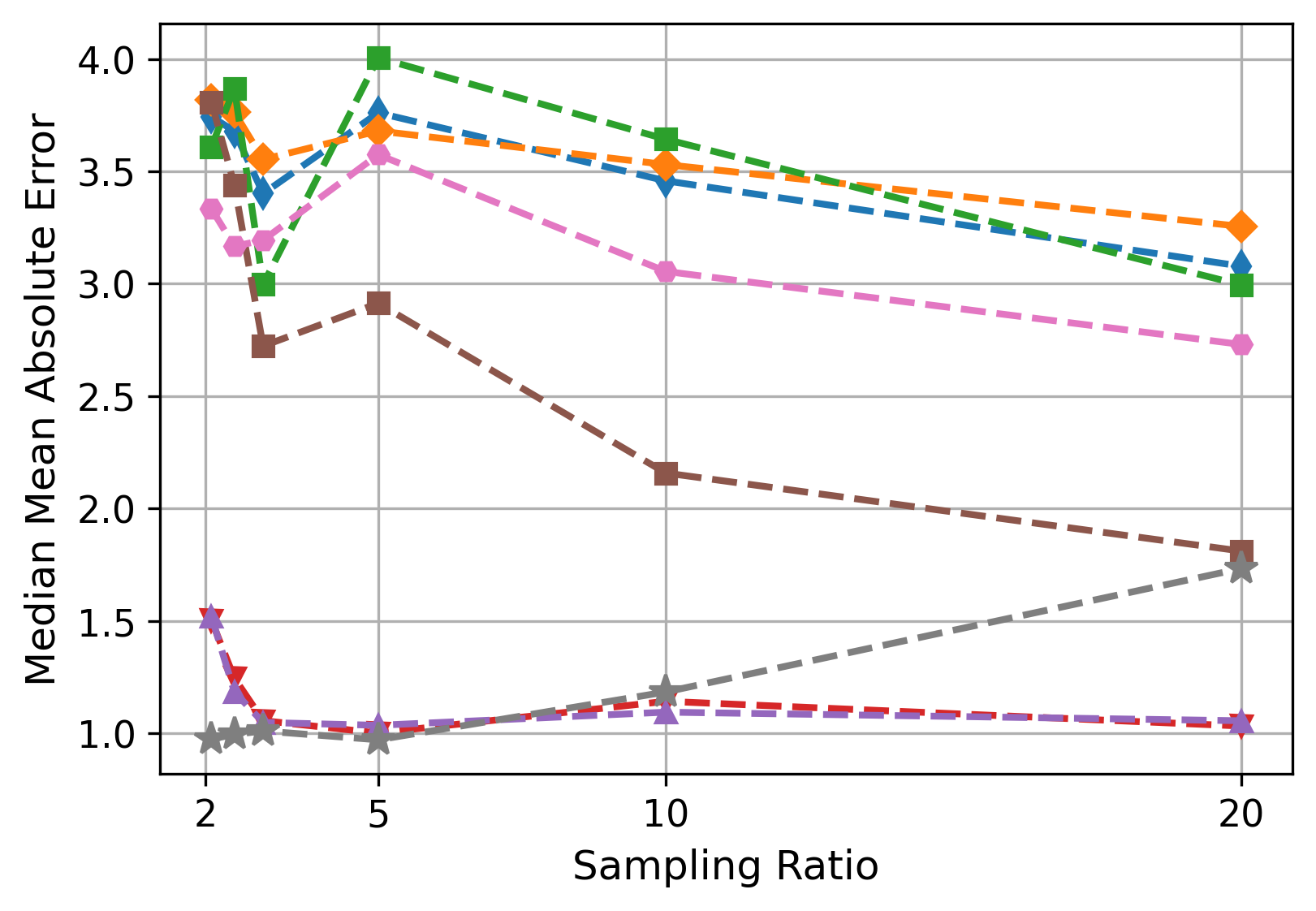} \\
        %(a) & (b) & (c) \\
        \includegraphics[width=0.3\textwidth]{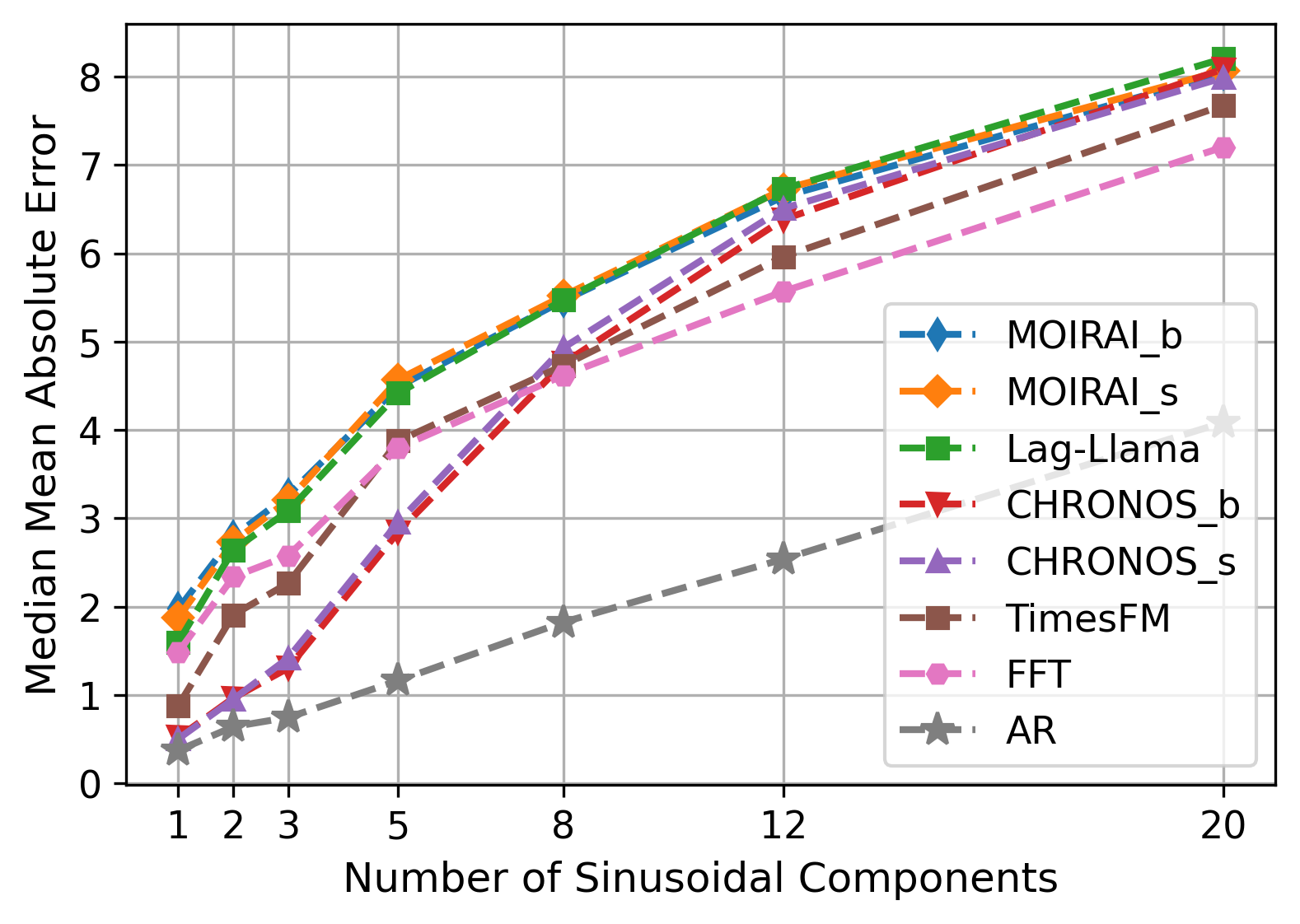} &
        \includegraphics[width=0.3\textwidth]{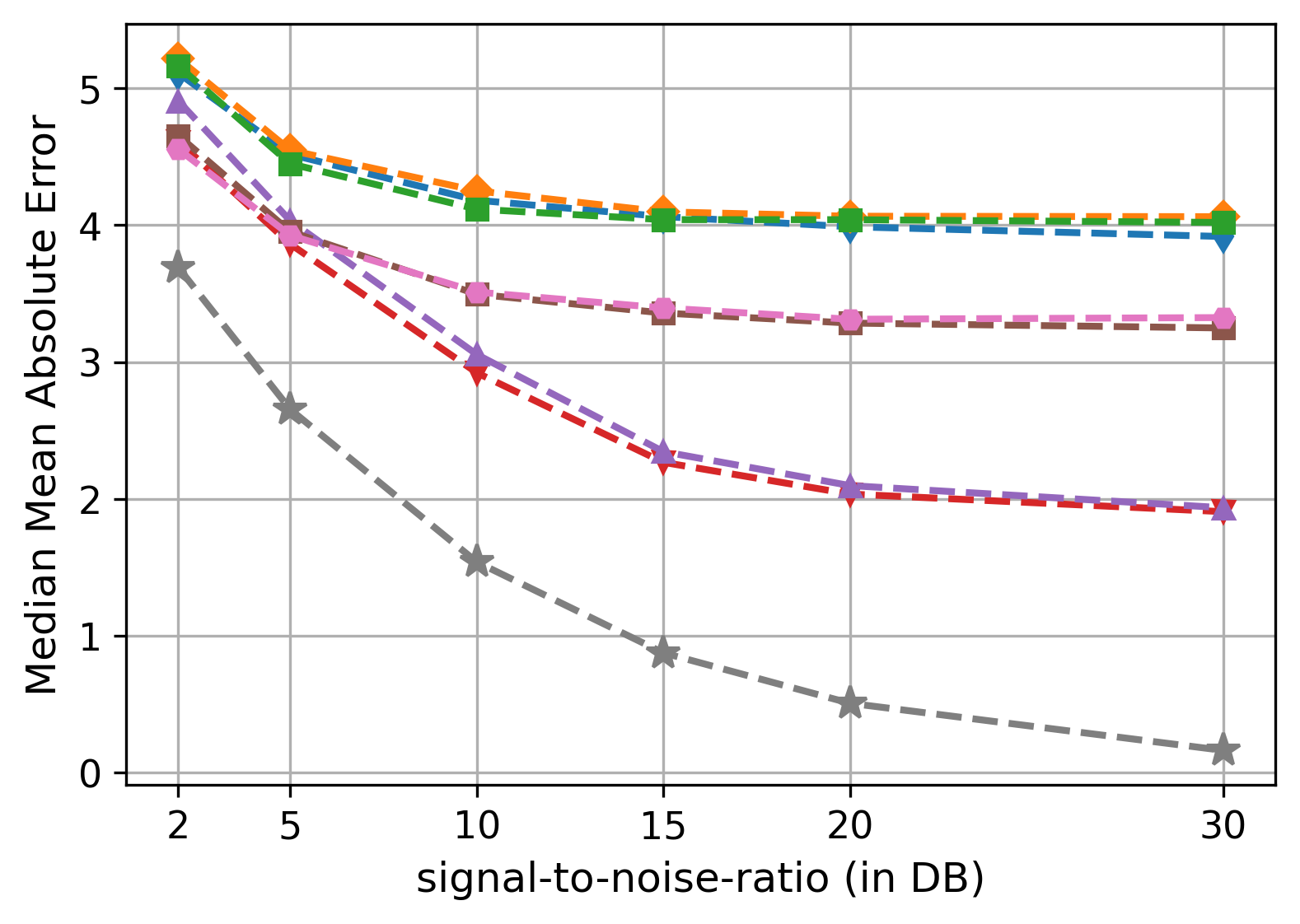} &
        \includegraphics[width=0.3\textwidth]{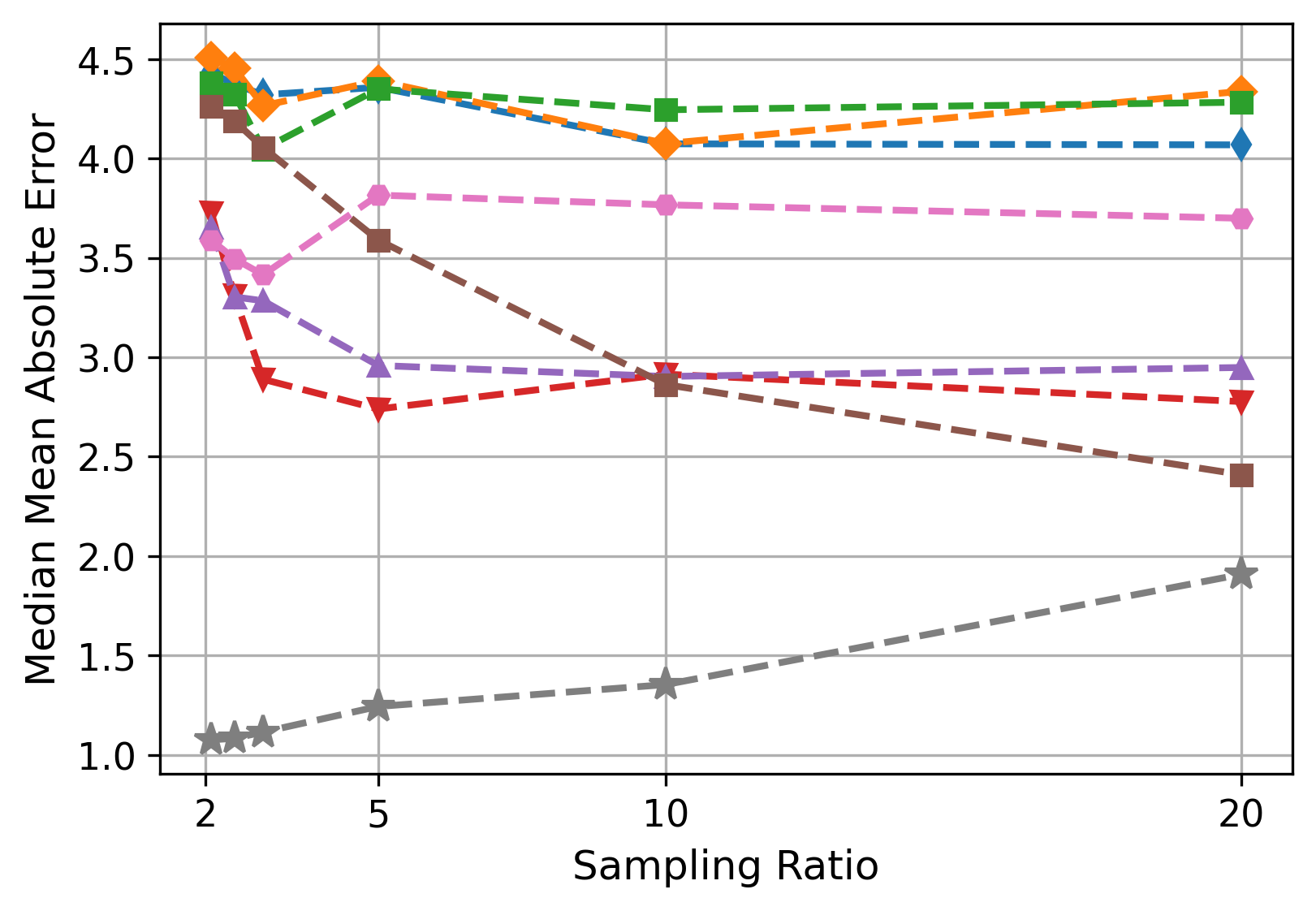} \\
       % (d) & (e) & (f)
    \end{tabular}
    \caption{Median Mean Absolute Error as a function of number of sinusoidal components, signal-to-noise ratio and sampling ratio. Top row represents Set A and bottom row represents Set B}
    \label{fig:MAE_median}
\end{figure}

\begin{figure}[ht!]
    \centering
    \begin{tabular}{cc}
        \includegraphics[width=0.45\textwidth]{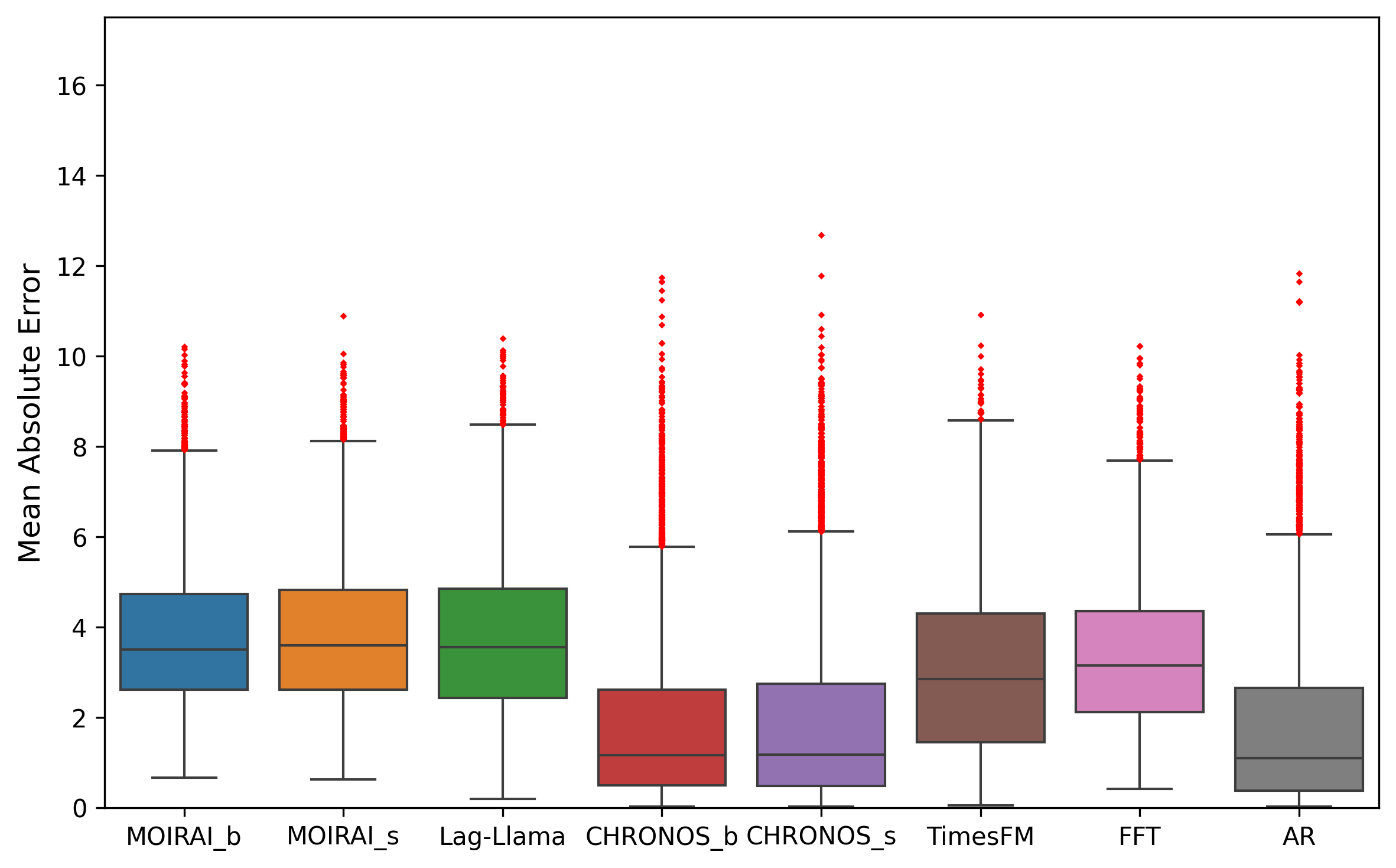} &
        \includegraphics[width=0.45\textwidth]{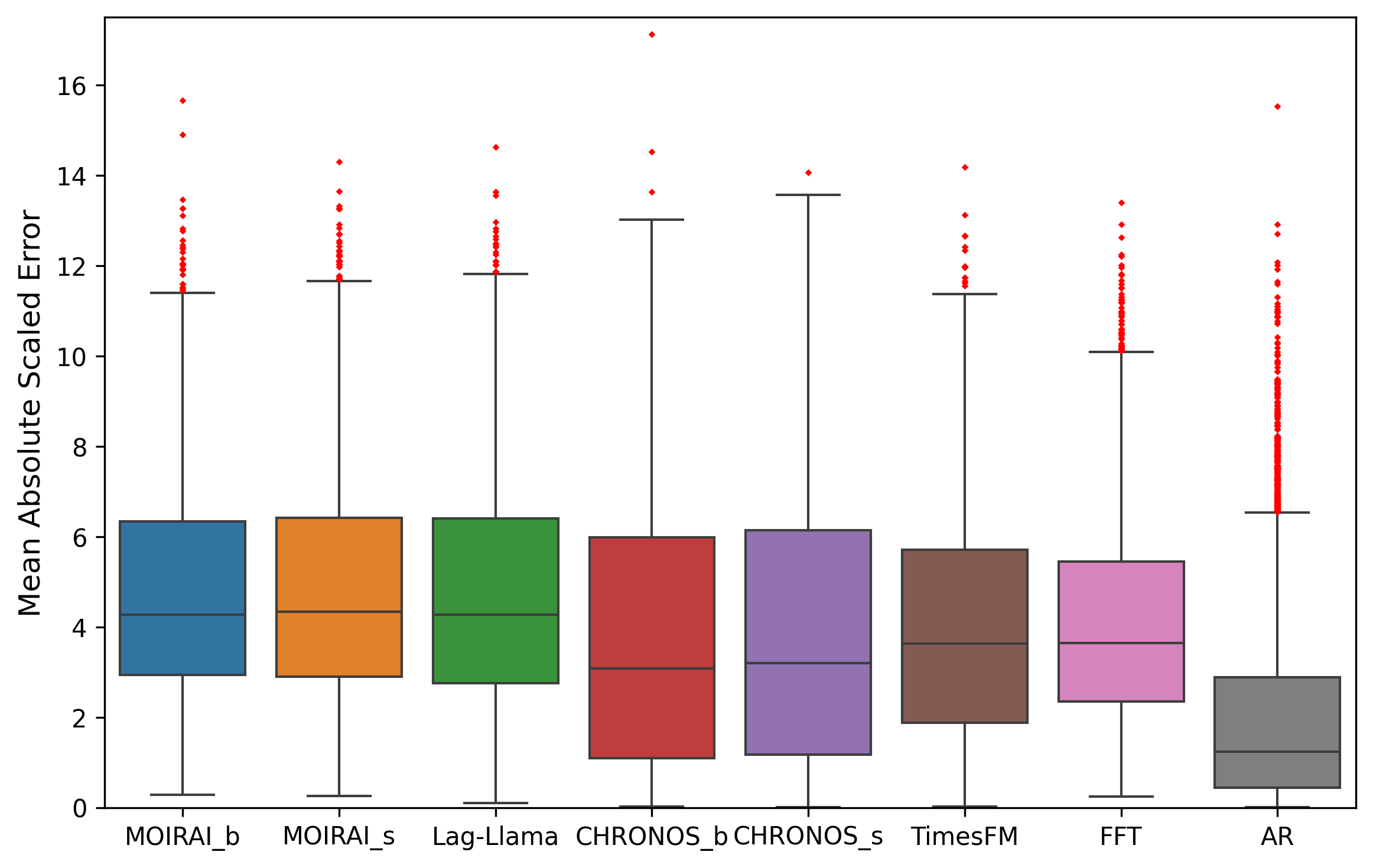} \\
        % (a) Caption for the first figure. & (b) Caption for the second figure. \\
    \end{tabular}
    \caption{Boxplots of mean absolute errors (lower the better), left: set A, right: set B}
    \label{fig:boxplot_mae}
\end{figure}

\begin{figure}[ht!]
    \centering
    \begin{tabular}{cc}
        \includegraphics[width=0.45\textwidth]{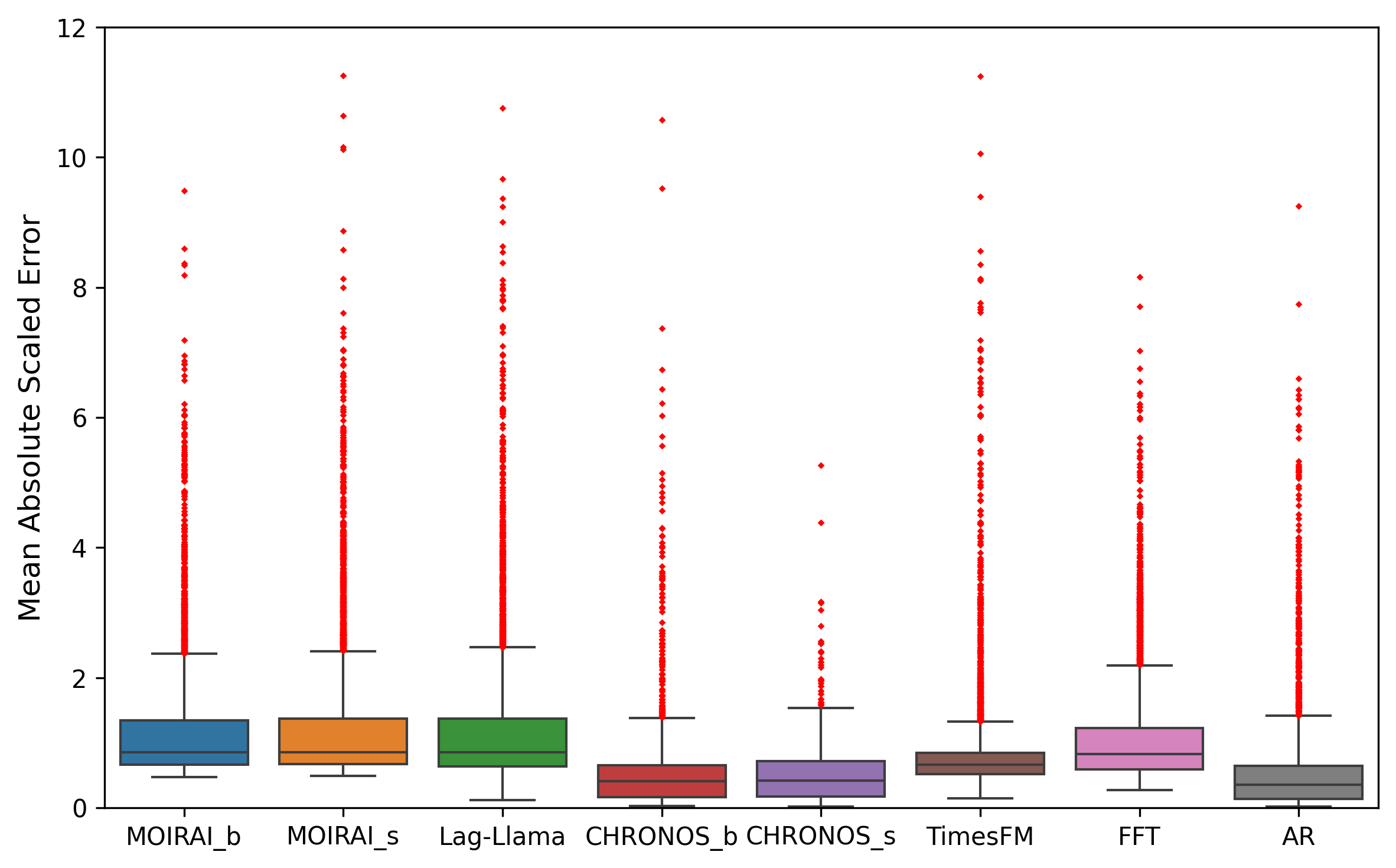} &
        \includegraphics[width=0.45\textwidth]{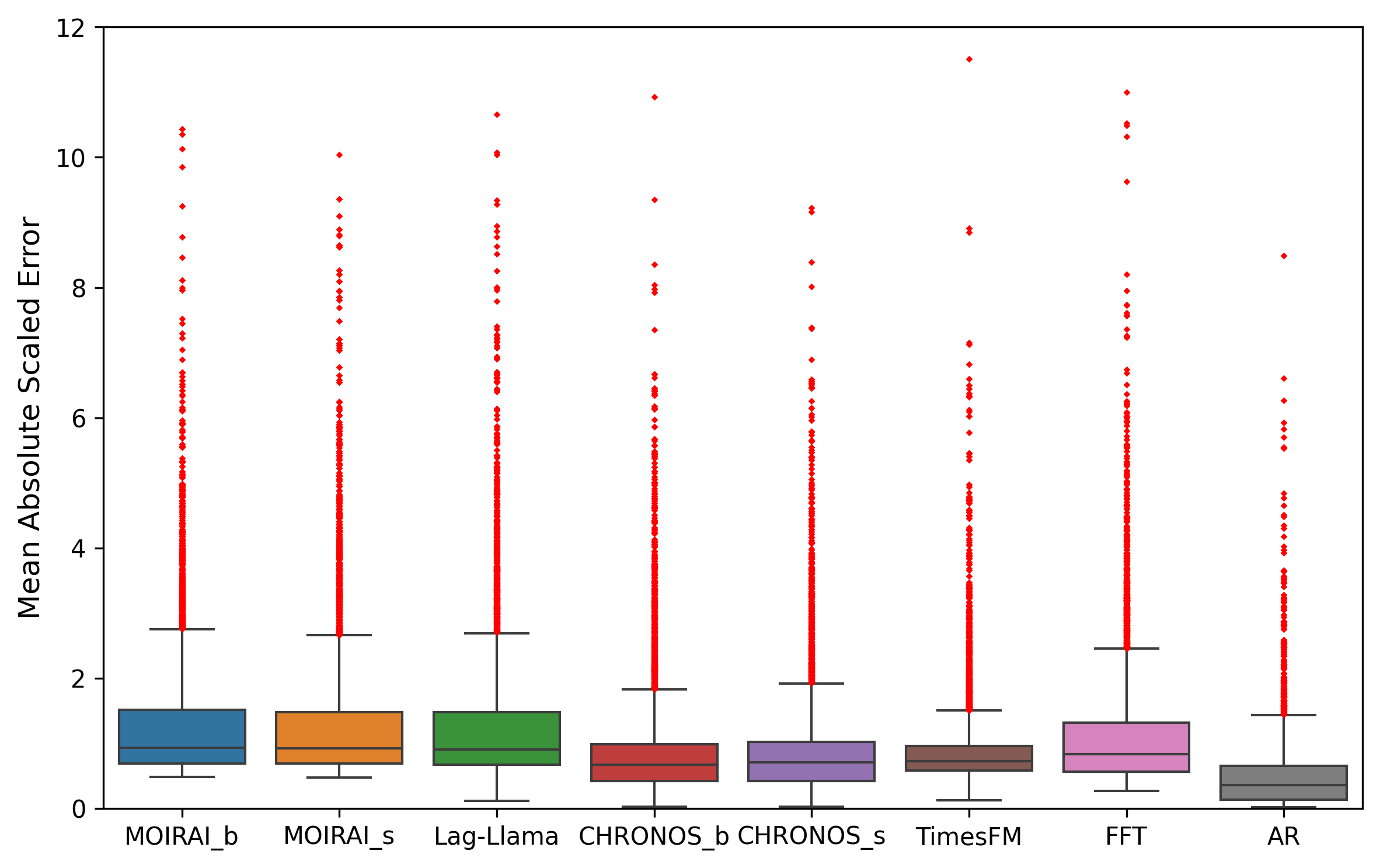} \\
        % (a) Caption for the first figure. & (b) Caption for the second figure. \\
    \end{tabular}
    \caption{Boxplots of mean absolute scaled errors (lower the better), left: set A, right: set B}
    \label{fig:boxplot_mase}
\end{figure}

\begin{figure}[ht!]
    \centering
    \begin{tabular}{cc}
        \includegraphics[width=0.45\textwidth]{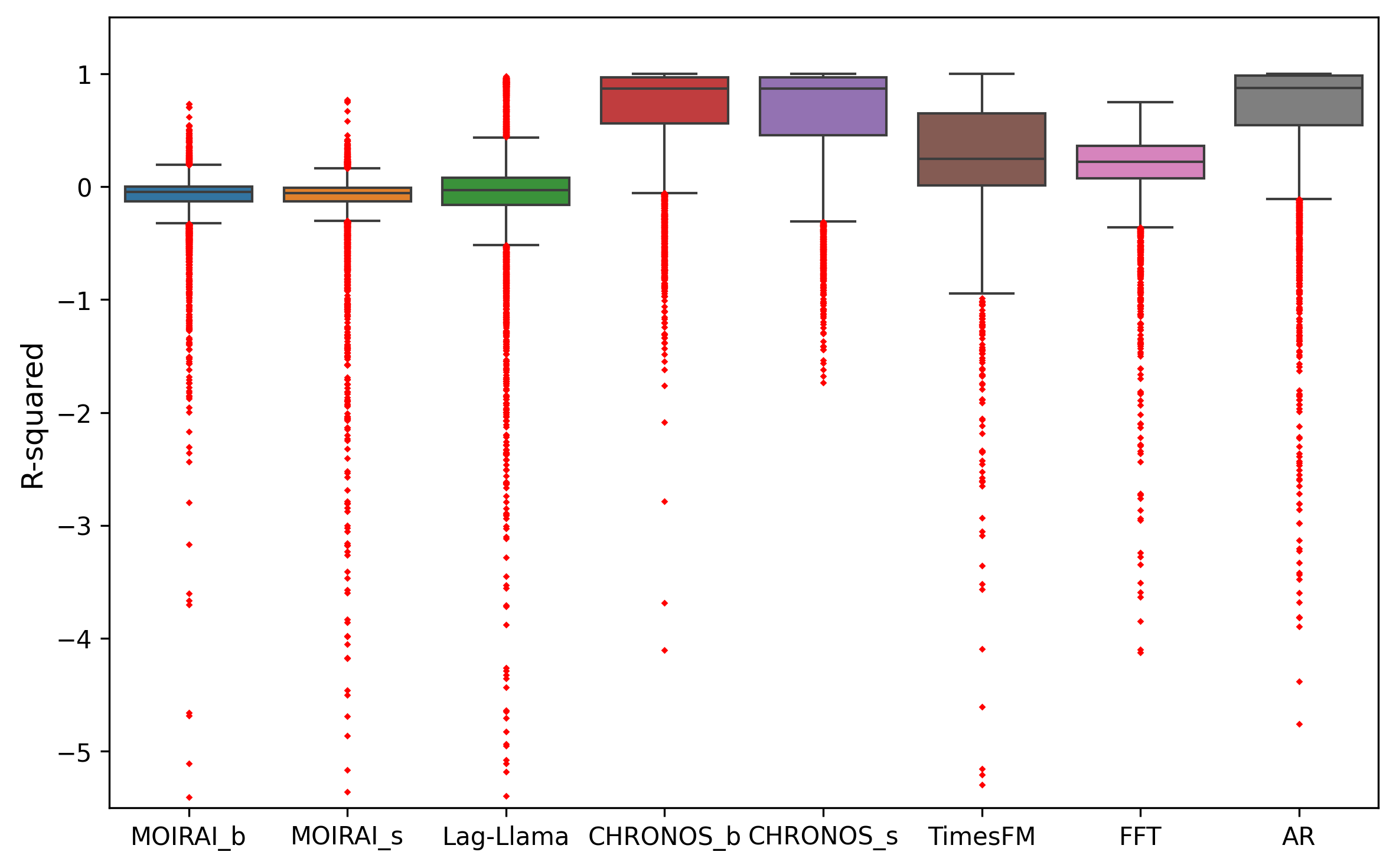} &
        \includegraphics[width=0.45\textwidth]{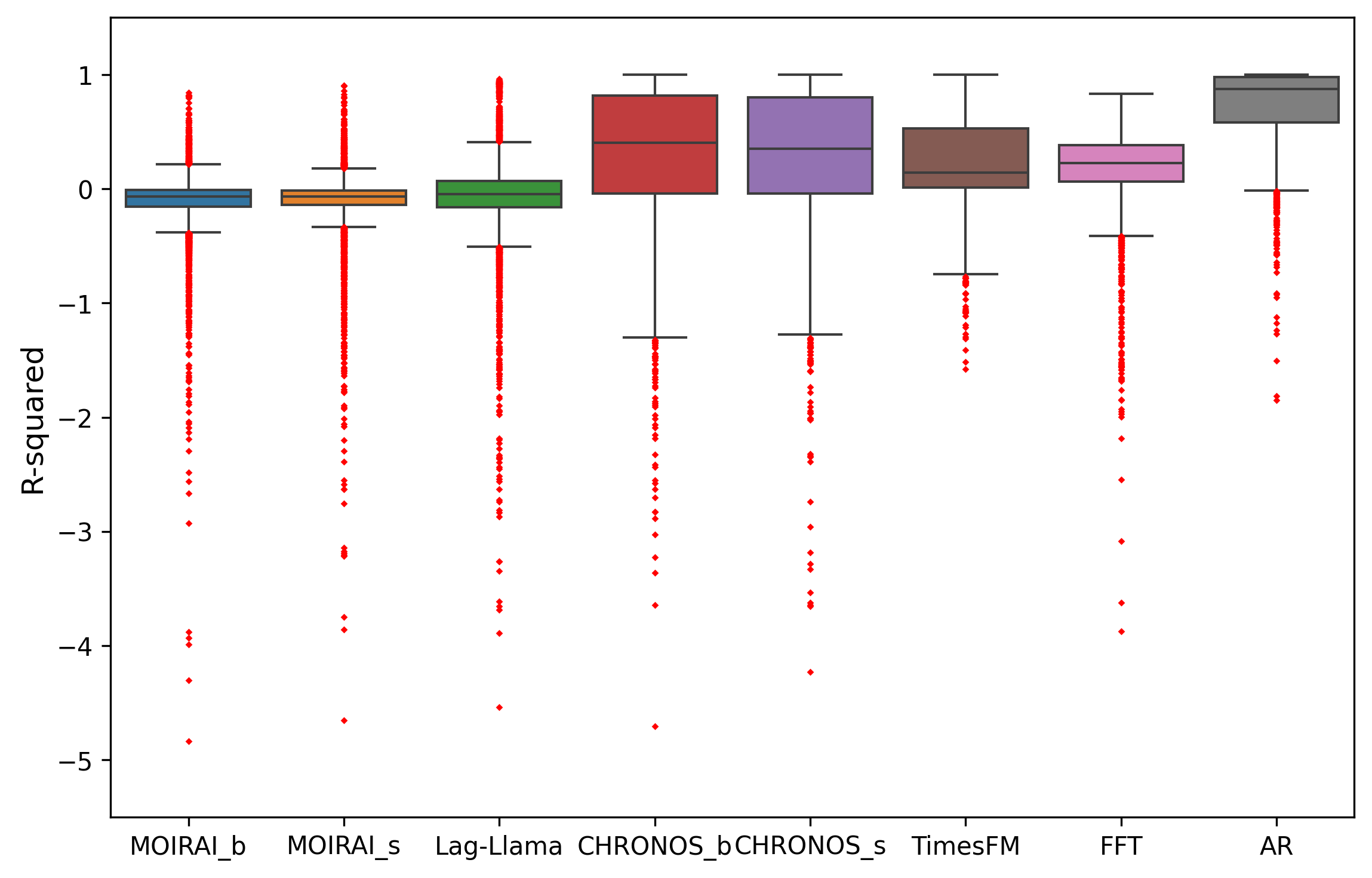} \\
        % (a) Caption for the first figure. & (b) Caption for the second figure. \\
    \end{tabular}
    \caption{Boxplots of R-squared (higher the better, maximum possible = 1.0), left: set A, right: set B}
    %\label{fig:boxplot_mae}
\end{figure}

% Optionally include extra information (complete proofs, additional experiments and plots) in the appendix.

\end{document}